\RequirePackage{rotating}
\RequirePackage{amsmath}
\documentclass[smallcondensed]{svjour3}
\usepackage[utf8]{inputenc}
\usepackage{amsmath}
\usepackage{amssymb}
\usepackage[rgb]{xcolor}
\usepackage{hyperref}
\usepackage{graphicx}
\usepackage{multirow}
\usepackage{mathtools}
\usepackage{hhline}
\usepackage{algorithm}
\usepackage{stmaryrd}
\usepackage{booktabs}
\usepackage{placeins}
\usepackage{adjustbox}
\usepackage[caption=false]{subfig}

\def\ptFiguresDirectory#1{./#1}
\def\ptRow2#1#2#3{{#1}_{#21},{#1}_{#22},\ldots,{#1}_{#2#3}}
\def\ptVec2#1#2#3{\left[\ptRow2{#1}{#2}{#3} \right]}
\def\ivBrack#1{\left\llbracket{#1}\right\rrbracket}
\newsavebox\CBox
\def\textBF#1{\sbox\CBox{#1}\resizebox{\wd\CBox}{\ht\CBox}{\textbf{#1}}}
\def\ptVecbfA#1#2{\left[{\textbf{#1}}_{1},{\textbf{#1}}_{2},\ldots,{\textbf{#1}}_{#2}\right]}
\newcommand{\argmax}[1]{\mathrm{argmax}_{{#1}}}
\def\argmaxX{\mathop{\arg\!\max}}

\def\vec#1{\boldsymbol{#1}}

\def\undefinedSymbol{\varnothing}

\title{ Dynamic classifier chains for multi-label learning}
\author{Pawel Trajdos \and Marek Kurzynski}
\authorrunning{P Trajdos, M Kurzynski}
\institute{Department of Systems and Computer Networks, Wroclaw
University of Technology, \\ Wybrzeze Wyspianskiego 27, 50-370
Wroclaw, Poland \\ \texttt{pawel.trajdos@pwr.wroc.pl}}

\begin{document}

\maketitle
\begin{abstract}
 In this paper, we deal with the task of building a dynamic ensemble of chain classifiers for multi-label classification. To do so, we proposed two concepts of classifier chains algorithms that are able to change label order of the chain without rebuilding the entire model. Such modes allows anticipating the instance-specific chain order without a significant increase in computational burden. The proposed chain models are built using the Naive Bayes classifier and nearest neighbour approach as a base single-label classifiers. To take the benefits of the proposed algorithms, we developed a simple heuristic that allows the system to find relatively good label order. The heuristic sort labels according to the label-specific classification quality gained during the validation phase. The heuristic tries to minimise the phenomenon of error propagation in the chain. The experimental results showed that the proposed model based on Naive Bayes classifier the above-mentioned heuristic is an efficient tool for building dynamic chain classifiers. 
\end{abstract}

\section{Introduction}\label{sec:Introduction}
Under well-known single-label classification framework, an object is assigned to only one class which provides a full description of the object. However, many real-world datasets contain objects that are assigned to different categories at the same time. All of these categories constitute a full description of the object. Omitting of one of these concepts induces a loss of information. Classification process in which such kind of data is involved is called multi-label classification~\cite{Gibaja2014}. A great example of a multi-label dataset is a gallery of tagged photos. Each photo may be described using such tags as mountains, sea, forest, beach, sunset, etc. Multi-label classification is a relatively new idea that is explored extensively for last two decades. As a consequence, it was employed in a wide range of practical applications including text classification~\cite{Jiang2012}, multimedia classification~\cite{Sanden2011} and bioinformatics~\cite{Wu2014} to name a few.

Multi-label classification algorithms can be broadly partitioned into two main groups i.e. dataset transformation algorithms and algorithm adaptation approaches~\cite{Gibaja2014}.

Methods belong to the group of algorithm adaptation approaches provides a generalisation of an existing multi-class algorithm. The generalised algorithm is able to solve multi-label classification problem in a direct way. Among the others, the most known approaches from this group are: multi label KNN algorithm~\cite{Jiang2012}, the Structured SVM approach~\cite{Diez2014} or deep-learning-based algorithms~\cite{Wei2015}.

In this paper, we investigate only dataset transformation algorithms that decompose a multi-label problem into a set of single-label classification tasks. To reconstruct a multi-label response, during the inference phase, outputs of the underlying single-label classifiers are combined in order to create a multi-label prediction.

Let's focus on one of the simplest decomposition methods. That is the \textit{binary relevance} (BR) approach that decomposes a multi-label classification task into a set of \textit{one-vs-rest} binary classification problems~\cite{AlvaresCherman2010}. This approach assumes that labels are conditionally independent. However the assumption does not hold in most of real-life recognition problems, the BR framework is one of the most widespread multi-label classification methods~\cite{Tsoumakas_Katakis_Vlahavas_2008}. This is due to its excellent scalability and acceptable classification quality~\cite{Luaces2012}.

To preserve scalability of BR systems, and provide a model of inter-label relations, Read et al.~\cite{Read2009,Read2011} provided us with the \textit{Classifier Chain} model (CC) which establish a linked chain of modified one-vs-rest binary classifiers. The modification consists of an extension of the input space of single-label classifiers along the chain sequence. To be more strict, for a given label sequence, the feature space of each classifier along the chain is extended with a set of binary variables corresponding to the labels that precede the given one. The model implies that, during the training phase, input space of given classifier is extended using the ground-truth labels extracted from the training set. During the inference step, due to lack of the ground-truth labels, we employ binary labels predicted by preceding classifiers. The inference is done in a greedy way that makes the best decision for each of considered labels. That is, the described approach passes along the chain, information allowing CC to take into account inter-label relations at the cost of allowing the label-prediction-errors to propagate along the chain~\cite{Read2011}. This way of performing classification induces a major drawback of the CC system. That is, the CC classifier uses a kind of greedy strategy during the inference phase. This design allows classification errors to propagate along the chain. As a consequence, the performance of a chain classifier strongly depends on chain configuration~\cite{Senge2013}. To overcome these effects, the authors suggested to generate an \textit{ensemble of chain classifiers} (ECC). The ensemble consists of classifiers trained using different label sequences~\cite{Read2009}. 

The originally proposed ECC ensemble uses randomly generated label orders. Additionally, each chain classifier is built using a resampled dataset. This approach provides an additional diversity into the ensemble classifier. This simple, yet effective approach allows improving the classification quality significantly in comparison to single chain classifier. However, the intuition says that there is still room for improvement when we employ a more data-driven approach.

Indeed, later research shows that the members of the ensemble may be chosen in such a way that provides further improvement of classification quality. 
That is, Read et al. proposed a strategy which uses Monte Carlo sampling to explore the label sequence space in order to find a classifier chain that offers the highest classification quality~\cite{Read2014}. Another approach was proposed by  Liu et al.~\cite{Liu2010}. They introduced a method that builds a model of inter-label relations as a \textit{directed acyclic graph} (DAG). The weights of the graph are calculated using confusion and support for each pair of labels. Then, the ensemble is generated using topological sorting of the graph. Chen et al.~\cite{Chen2016} proposed a method that makes clusters of labels. Then, for each cluster of labels, an undirected graph describing inter-label relations is built. Then, a \textit{minimum spanning tree} is created for the graph. After that, \textit{ breadth-first search} algorithm determines sequences for a cluster-specific ensemble of CC classifiers.  A similar approach was proposed by Huang et al~\cite{Huang2015}. They proposed to build the clusters using a meta-space that mixes input space and label space. Then inter-label relations are modeled using correlation. The model is expressed using DAG structure. Finally, the CC classifier is built for each cluster. The chain structure may also be induced using Bayesian Network approach~\cite{Zhang2014}.   

Chain sequence can be also found using meta-heuristic approach. That is, Goncalves et al. developed a strategy that utilises a \textit{genetic algorithm} (GA) to find a good chain structure for the entire dataset~\cite{Goncalves2013,Gonalves2015}. The proposed approach using wrapper-based approach. That is each chromosome codes different chain order. To evaluate those label orders each corresponding classifier must be built and evaluated using a validation set. A similar approach was also used by Trajdos and Kurzynski who proposed to use a multi-objective genetic algorithm to optimize classification quality and chain diversity simultaneously~\cite{Trajdos2017}.   Although those methods are rather time-consuming, they provide a significant improvement in terms of classification quality.

Another way of dealing with the error propagation is to build a classifier that combines CC algorithm BR-based approach~\cite{Montaes2014}. The authors proposed a stacking based architecture to combine the above-mentioned classifiers. That is, the first layer is a simple BR classifier that predicts each label separately. The attribute set of the classifiers from the second layer is extended using all labels except the predicted one.  During the training phase, both layers are trained separately. During the prediction phase, on the other hand,  classifiers from the second layer mix outcomes of the BR classifiers with the outcomes provides by preceding classifiers. That is, the first classifier of the chain structure has its attribute space extended by the outcomes of the BR classifier. The second one uses the prediction of the first one and the remaining attributes are taken from the prediction of the BR classifier. Finally, the last classifier along the chain has the attribute space extended using only labels predicted by the preceding classifiers. Another way of combining the CC classifier with the BR classifier is described in~\cite{Madjarov2012}.

The previously cited methods build ensemble structure during the training procedure. Consequently, throughout this paper, this kind of methods will be called static methods. The dynamic chain classifiers, on the other hand, determines the best label order at the prediction phase~\cite{daSilva2014}. The above-mentioned classifier produces a set of randomly generated label sequences and then validates the chain classifiers. During the validation phase, each point from the validation set is assigned with a label order that produces the most accurate output vector for this point. As the experimental research shows, the dynamic methods of building a label order may achieve better classification quality~\cite{daSilva2014}.   

We observed that during the building of a dynamic chain classifier, multiple chain classifiers must be learned. These classifiers are built using the same training set and differ only in chain order.  As a consequence, the computational burden of the algorithm may be reduced if there exists a classifier that is trained once and changing the label sequence is done without rebuilding the model. To address this issue, we built two models based on the Naive Bayes~\cite{Hand2001} approach and the nearest neighbour approach~\cite{Cover1967} that meet the above-mentioned properties. 

Additionally, we proposed a dynamic method of determining the chain order based on classification quality for each label separately.

A part of this paper was previously published in~\cite{Trajdos2017b}. This paper is an extended version of the previously-published work. The main elements that has been changed/extended:
\begin{itemize}
 \item The literature review has been extended. 
 \item We have added the results of the experimental comparison of the BR and CC versions of different base classifiers.
 \item We have proposed a new model of the dynamic chain classifier. That is, we introduce the CC model based on the nearest neighbour approach. 
 \item New experimental results have been provided.
\end{itemize}

The rest of the paper is organised as follows. Next Section~\ref{sec:ProposedMeth} provides a formal description of the multi-label classification problem and describes the developed algorithms. Section~\ref{sec:ExpSetup} contains a description of the conducted experiments. The results are presented and discussed in Section~\ref{sec:ResDisc}. Finally, Section~\ref{sec:Conclusions} concludes the paper. 

\section{Proposed Methods}\label{sec:ProposedMeth}

In this section, we introduce a formal notation of multi-label classification problem and provide a description of the proposed method.
\subsection{Preliminaries}\label{subsec:Preliminaries}
Under the \textit{multi-label} (ML) formalism a $d-\mathrm{dimensional}$ object $\vec{x}=\ptVec2{x}{}{d}\in\mathcal{X}$ is assigned to a set of labels indicated by a binary vector of length $L$: $\vec{y}=\ptVec2{y}{}{L}\in\mathcal{Y}=\{0,1\}^{L}$, where $L$ denotes the number of labels.

In this paper, we follow a statistical classification framework. As a consequence, it is assumed that object $\vec{x}$ and its set of labels $y$ are realizations of corresponding random vectors ${\vec{\textbf{X}}=\ptVecbfA{X}{d} }$, ${\vec{\textbf{Y}}=\ptVecbfA{Y}{L}}$ and the joint probability distribution $P(\vec{\textbf{X}},\vec{\textbf{Y}})$ on $\mathcal{X}\times\mathcal{Y}$ is known.

Because the above-mentioned assumption is never meet in real world, in this study, we assume that  multi-label classifier $h: \mathcal{X} \mapsto \mathcal{Y}$, which  maps feature space $\mathcal{X}$ to the set $\mathcal{Y}$, is built in a supervised learning procedure using the training set $\mathcal{T}$ containing $N$ pairs of feature vectors $\vec{x}$ and corresponding class labels $\vec{y}$:
\begin{equation}\label{eq:trainSet} 
\mathcal{T}=\left\{(\vec{x}^{1},\vec{y}^{1}), (\vec{x}^{2},\vec{y}^{2}), \ldots ,(\vec{x}^{N},\vec{y}^{N})\right\}.
\end{equation}

\subsection{Naive Bayes Classifier for Dynamic Classifier Chains}\label{subsec:NBCC}
In this paper, we consider ML classifiers build according to the chain rule. That is, the classifier $h$ is an ensemble of $L$ single-label classifiers $\psi_i$ that constitutes a linked chain which is built according to a permutation of label sequence $\pi$. As it was mentioned earlier, in this paper we follow the statistical classification framework. Consequently, each single-label classifier $h_{\pi(i)}$ along with the chain makes its decision according to the following rule:
\begin{equation}\label{eq:chainDecisionRule}
	h_{\pi(i)}(\vec{x}) =\argmax{y\in \{0,1\}}P(\textbf{Y}_{\pi(i)}=y|B_{\pi(i)}(\vec{x})),
\end{equation}
where $B_{\pi(i)}(\vec{x})$ is a random event defined below:
\begin{align}\label{eq:ChainEvent}
	B_{\pi(i)}(\vec{x}) &= (\vec{\textbf{X}}=\vec{x},\textbf{Y}_{\pi(i-1)}=h_{\pi(i-1)}(\vec{x})\textbf{Y}_{\pi(i-2)}=h_{\pi(i-2)}(\vec{x}),\cdots\nonumber\\
	&\,, \textbf{Y}_{\pi(1)}=h_{\pi(1)}(\vec{x})),\forall i \in \left\{2,3,\cdots,L\right\},
\end{align}
and for $i=1$
\begin{align}\label{eq:ChainEvent2}
 B_{\pi(1)}(\vec{x}) &= \left(\vec{\textbf{X}}=\vec{x}\right).
\end{align}
Conditioning on the random event $B_{\pi(i)}(\vec{x})$ instead of $\vec{\textbf{X}}=\vec{x}$ allows the chain to take inter-label dependencies into account. 
The above-mentioned classification rule is a greedy rule that calculates the probability~\eqref{eq:chainDecisionRule} using predictions of preceding classifiers. The optimal prediction under the chaining rule may be found using the PCC approach~\cite{cheng2010bayes}. However, the approach requires the number of calculations that grows exponentially with the number of labels.

The probability defined in~\eqref{eq:chainDecisionRule} is then computed using the Bayes rule:
\begin{align}\label{eq:DecRuleProbBayes}
P(\textbf{Y}_{\pi(i)}=y|B_{\pi(i)}(\vec{x}))  &= \dfrac{P(\textbf{Y}_{\pi(i)}=y)}{P(B_{\pi(i)}(\vec{x}))}P(B_{\pi(i)}(\vec{x})|\textbf{Y}_{\pi(i)}=y).
\end{align}
The term $P(B_{\pi(i)}(\vec{x}))$ does not depend on event $\textbf{Y}_{\pi(i)}=y$. Consequently, the decision rule~\eqref{eq:chainDecisionRule} is rewritten:
\begin{align}\label{eq:ChainDecisionRuleSimple}
h_{\pi(i)}(\vec{x}) &=\argmax{y\in \{0,1\}}P(\textbf{Y}_{\pi(i)}=y)P(B_{\pi(i)}(\vec{x})|\textbf{Y}_{\pi(i)}=y)
\end{align}

Now, to improve the readability we simplify the notation:
\begin{equation}
P(B_{\pi(i)}(\vec{x})|\textbf{Y}_{\pi(i)}=y) = P(B_{\pi(i)}(\vec{x})|y).
\end{equation}
Then, following the Naive Bayes rule, we assume that all random variables that constitute $B_{\pi(i)}(\vec{x})$ are conditionally independent given $\textbf{Y}_{\pi(i)}=y$. Consequently,  $P(B_{\pi(i)}(\vec{x})|y)$ is defined using the following formula:
\begin{align}\label{eq:ChainNBFormula}
	P(B_{\pi(i)}(\vec{x})|y) &= \prod_{m=1}^{d}P(\vec{\textbf{X}}_{m}=\vec{x}_m|y)\prod_{{l=1}}^{l=i-1}P(\textbf{Y}_{\pi(l)}=h_{\pi(l)}(\vec{x})|y).
\end{align}

Now, it is easy to see that the term $\prod_{{l=1}}^{l=i-1}P(\textbf{Y}_{\pi(l)}=h_{\pi(l)}(\vec{x})|y)$, contrary to $\prod_{m=1}^{d}P(\vec{\textbf{X}}_{m}=\vec{x}_m|y)$, depends on the chain structure. 
Furthermore, all probability distributions used in the above-mentioned terms can be estimated during the training phase when the chain structure is unknown.

The training and inference phases are described in detail using pseudocode shown in Algorithms~\ref{fig:TrainNB} and~\ref{fig:InferenceNB}.
{
\begin{algorithm}[tb]
\vbox{%
\begin{center}
\caption{Pseudocode of the learning procedure of the Naive Bayes classifier.}
\label{fig:TrainNB}%
\tt
\smallskip
\scriptsize
\begin{tabbing}

\quad \=\quad \=\quad \=\quad \=\quad \=\quad \kill
\textbf{Input data:}\\
\>$\mathcal{T}$ - training set;\\

\textbf{BEGIN}\\
\>Split $\mathcal{T}$ into $\mathcal{T}_A$ and $\mathcal{V}$ so that:\\
\> $|\mathcal{T}_A| = t|\mathcal{T}|$ and $|\mathcal{V}| = (1-t)|\mathcal{T}|$, $t \in (0,1)$\\
\> $\mathcal{T}_A \cap \mathcal{V} = \emptyset$;\\
\>Using $\mathcal{T}_A$ build estimators of\\
\> the following distributions:\\
\>\>$P(\textbf{Y}_{\pi(i)}=y)\forall i\in \{1,2,\cdots,L\}, y\in \{0,1\}$\\
\>\>$P(\vec{\textbf{X}}_{m}|\textbf{Y}_{\pi(i)}=y)\forall i\in \{1,2,\cdots,L\}, y\in \{0,1\}, m\in \{1,2,\cdots,d \}$\\
\>\>$P(\textbf{Y}_{\pi(l)}|\textbf{Y}_{\pi(i)}=y)\forall i,l\in \{1,2,\cdots,L\};i\neq l$\\
\textbf{END}
\end{tabbing}
\end{center}
}
\end{algorithm}
}

\begin{algorithm}[tb]
\vbox{%
\begin{center}
\caption{Pseudocode of the inference procedure  of the Naive Bayes classifier.}
\label{fig:InferenceNB}%
\tt
\smallskip
\scriptsize
\begin{tabbing}
\quad \=\quad \=\quad \=\quad \=\quad \=\quad \kill
\textbf{Input data:}\\
\>$\vec{x}\in\mathcal{X}$ -- input instance;\\
\>$\mathcal{V}$ -- validation set;\\

\textbf{BEGIN}\\
\>\#Query the BR models\\
\>\textbf{FOR} $i\in \{1,2,\cdots,L\}$:\\
\>\>$e_{i}^{0} = \prod_{m=1}^{d}P(\vec{\textbf{X}}_{k}=\vec{x}_m|\textbf{Y}_{i}=0)$;\\
\>\>$e_{i}^{1} = \prod_{m=1}^{d}P(\vec{\textbf{X}}_{k}=\vec{x}_m|\textbf{Y}_{i}=1)$;\\
\>\textbf{END FOR;}\\
\>Determine label permutaion $\pi$ using $\mathcal{V}$ and $\vec{x}$;\\
\> \textbf{SET} $i=1$;\\
\>\textbf{DO:}\\
\>\>$h_{\pi(i)}(\vec{x}) = \argmaxX_{y\in \{0,1\}} e_{\pi{(i)}}^yP(\textbf{Y}_{\pi(i)}=y)$\\
\>\>\textbf{FOR} $j\in\{i+1,i+2,\cdots,L\}$:\\
\>\>\>$d_{\pi{(j)}}^{0} := e_{\pi{(j)}}^{0} * P(\textbf{Y}_{\pi(i)}=h_{\pi(i)}(\vec{x})|\textbf{Y}_{\pi(j)}=0)$\\
\>\>\>$d_{\pi{(j)}}^{1} := e_{\pi{(j)}}^{1} * P(\textbf{Y}_{\pi(i)}=h_{\pi(i)}(\vec{x})|\textbf{Y}_{\pi(j)}=1)$\\
\>\>\textbf{END FOR;}\\
\>\>$i:=i+1;$\\
\>\textbf{WHILE}($i<L$);\\
\>\textbf{RETURN} $[h_{1}(\vec{x}),h_{2}(\vec{x}),\cdots,h_{L}(\vec{x})]$;\\
\textbf{END}
\end{tabbing}
\end{center}
}
\end{algorithm}

\subsubsection{Computational complexity}\label{subsec:CompComplex}

In this section, we assess the increase in computational complexity that the proposed algorithm causes. 

First of all, it is easy to see that for both the original and the proposed algorithm the number of estimators that must be built to assess $P(\vec{\textbf{X}}_{m}|\textbf{Y}_{\pi(i)}=y)\forall i\in \{1,2,\cdots,L\}, y\in \{0,1\}$ is: $2Ld$.

The number of estimators of $P(\textbf{Y}_{\pi(i)}=y)\forall i\in \{1,2,\cdots,L\}, y\in \{0,1\}$ that must be built is also the same for both classifiers: L.

The key difference is in the number of estimators of $P(\textbf{Y}_{\pi(l)}|\textbf{Y}_{\pi(i)}=y)$ that must be built. For the original CC classifier the number of estimators that is built is $L(L-1)$. On the other hand our method builds $2L^{2}$ estimators. 

At the inference phase, the only additional calculations are performed to determine the permutation of labels. Since the validation set is involved in this process, a number of calculations is proportional to $O(|\mathcal{V}|L)$. 

\subsection{KNN Classifier for Dynamic Classifier Chains}\label{subsec:KNNClass}

In this section, we define a dynamic classifier chain algorithm based on the nearest neighbours approach.The nearest neighbour algorithm is an instance-based classifier that does not build an explicit model of mapping between the feature space and the label space. Instead, the classifier performs classification in a lazy manner. That is, the R nearest instances and then the class is predicted using labels of the neighbour instances. 

Let's begin with the definition of a distance function that depends on label permutation and the position along the chain. The distance function is defined in the extended feature space that combines the input space and the label space. For the first position, the distance is a simple Euclidean distance in the input space: 
\begin{align}\label{eq:knnPermDepDist1}
 \delta_{\pi,i=1}\left( \left( \vec{x},\vec{y} \right),\left( \vec{x}^{\prime},\vec{y}^{\prime} \right)\right) &= \sqrt{\sum_{j=1}^{d}\left( \vec{x}_j - \vec{x}_{j}^{\prime} \right)^{2}}.
\end{align}
For the other positions, the distance function uses both the input space and the label space: 
\begin{align}\label{eq:knnPermDepDist2}
 \delta_{\pi,i}\left( \left( \vec{x},\vec{y} \right),\left( \vec{x}^{\prime},\vec{y}^{\prime} \right)\right) &= \sqrt{{ \sum_{j=1}^{d}\left( \vec{x}_j - \vec{x}_j^{\prime} \right)^{2}}{ + \sum_{l=1}^{i-1}(\vec{y}_{\pi(l)} - \vec{y}^{\prime}_{\pi(l)})^2}}.
\end{align}
Such defined distance function allows us to make the prediction using chaining rule. Since the distance is modified in order to fit the chain structure. During the inference phase, the distance calculates the extended distance using labels predicted at the preceding steps of the procedure. 
The above-defined distance function is used to build the neighbourhood of a given point in the extended feature space: $M_{\pi,i}^{R}((\vec{x},\vec{y}))$. The neighbourhood contains the $R$ closest instances selected from the training set according to the distance function $\delta_{\pi,i}$.

Given the neighbourhood, the probability $P(\textbf{Y}_{\pi(i)}=y|B_{\pi(i)}(\vec{x}))$ is estimated as follows: 
\begin{align}\label{eq:knnProbability Approx}
 P(\textbf{Y}_{\pi(i)}=y|B_{\pi(i)}(\vec{x})) &\approx \dfrac{\left|\left\{ {\left( \vec{x}^{n},\vec{y}^{n} \right) |}{ \left( \vec{x}^{n},\vec{y}^{n} \right)\in M_{\pi,i}^{R}((\vec{x},\vec{h}(\vec{x}))), \, \vec{y}^{n}_{i} = y}   \right\}\right|}{R}.
\end{align}

The label $\pi(i)$ is also predicted using rule~\eqref{eq:chainDecisionRule}.

{
\begin{algorithm}[tb]
\vbox{%
\begin{center}
\caption{Pseudocode of the learning procedure of the nearest neighbour classifier.}
\label{fig:TrainKNN}%
\tt
\smallskip
\scriptsize
\begin{tabbing}

\quad \=\quad \=\quad \=\quad \=\quad \=\quad \kill
\textbf{Input data:}\\
\>$\mathcal{T}$ - training set;\\

\textbf{BEGIN}\\
\>Split $\mathcal{T}$ into $\mathcal{T}_A$ and $\mathcal{V}$ so that:\\
\>\> $|\mathcal{T}_A| = t|\mathcal{T}|$ and $|\mathcal{V}| = (1-t)|\mathcal{T}|$, $t \in (0,1)$\\
\>\> $\mathcal{T}_A \cap \mathcal{V} = \emptyset$;\\
\>Save the training set $\mathcal{T}_A$\\
\textbf{END}
\end{tabbing}
\end{center}
}
\end{algorithm}
}

\begin{algorithm}[tb]
\vbox{%
\begin{center}
\caption{Pseudocode of the inference procedure  of the nearest neighbour classifier.}
\label{fig:InferenceKNN}%
\tt
\smallskip
\scriptsize
\begin{tabbing}
\quad \=\quad \=\quad \=\quad \=\quad \=\quad \kill
\textbf{Input data:}\\
\>$\vec{x}\in\mathcal{X}$ -- input instance;\\
\>$\mathcal{V}$ -- validation set;\\
\>$\mathcal{T}_A$ -- Training set;\\

\textbf{BEGIN}\\
\>\textbf{SET: }$\vec{h}(\vec{x}) = [h_{1}(\vec{x})=\undefinedSymbol,h_{2}(\vec{x})=\undefinedSymbol,\cdots,h_{L}(\vec{x})=\undefinedSymbol]$;\\
\>Determine label permutaion $\pi$ using $\mathcal{V}$ and $\vec{x}$;\\
\>\textbf{For $i\in \{1,2,\cdots,L \}$}\\
\>\textbf{BEGIN}\\
\>\>$h_{\pi(i)}(\vec{x}) = \argmaxX\limits_{{y\in\{0,1\}}}\dfrac{\left|\left\{\left( \vec{x}^{n},\vec{y}^{n} \right) | \left( \vec{x}^{n},\vec{y}^{n} \right)\in M_{\pi,i}^{R}((\vec{x},\vec{h}(\vec{x}))), \, \vec{y}^{n}_{i} = y    \right\}\right|}{R}$ \\
\>\textbf{END}\\
\>\textbf{RETURN} $[h_{1}(\vec{x}),h_{2}(\vec{x}),\cdots,h_{L}(\vec{x})]$;\\
\textbf{END}
\end{tabbing}
\end{center}
}
\end{algorithm}

The training procedure is described in Algorithm~\ref{fig:TrainKNN}. The procedure is very simple. That is, it splits the original training set into actual training set $\mathcal{T}_A$ and the validation set $\mathcal{V}$.

The inference procedure begins with assigning undefined values $\undefinedSymbol$ into the prediction vector $\vec{h}(\vec{x})$. Then the predictions are updated sequentially according to the permutation $\pi$. The precodure is shown in Algorithm~\ref{fig:InferenceKNN}

\subsection{Dynamic Chain order}\label{subsec:DynamicCC}
In this subsection, we define a local measure of classification quality. To do so, we employed a modified version of the well-known $F_1$ measure.

First of all, we defined a fuzzy neighbourhood in the input space. The neighbourhood of an instance $\vec{x}$ is defined using the following fuzzy set~\cite{Zadeh1965}:
\begin{align}\label{eq:neighSet} 
\mathcal{N}(\vec{x}) &= \left\{{ \left(\vec{x}^{n},\vec{y}^{n},\mu(\vec{x},\vec{x}^{n})\right):}{\left(\vec{x}^{n},\vec{y}^{n}\right)\in \mathcal{V}}\right\},
\end{align}
where each tripplet $\left(\vec{x}^{n},\vec{y}^{n},\zeta)\right)$ defines fuzzy set with the membership coefficient $\zeta$. 
The membership function $\mu(\vec{x},\vec{x}^{n})$ is defined using gaussian potential function:
\begin{equation}
 \mu(\vec{x},\vec{x}^{n}) =    \exp({-\beta \delta(\vec{x},\vec{x}^{n})^2}).
\end{equation}
The distance function $\delta(\vec{x},\vec{x}^{n})$ is simple euclidean distance and the $\beta$ coefficient is tuned during the experiments. 

Then, we define set of points that belongs to given label $\mathcal{V}_l$ and that are classified as given label $\mathcal{D}_l$:
\begin{align}
 \mathcal{V}_{l} &= \left\{ \left(\vec{x}^{n},\vec{y}^{n},1\right):\right.  \left. \left(\vec{x}^{n},\vec{y}^{n}\right)\in \mathcal{V},\vec{y}^{n}_{l} =1 \right\}\label{eq:oneInd}  \\
 \mathcal{D}_{l} &= \left\{  \left(\vec{x}^{n},\vec{y}^{n},1\right):\right.\left. \left(\vec{x}^{n},\vec{y}^{n}\right)\in \mathcal{V}, h^{BR}_{l}(\vec{x}^{n})=1 \right\}\label{eq:decSet}
\end{align}
The above-mentioned classifier responses are related to the binary relevance classifier that can be built without knowing the order of the chain. The classifier is defined using the following classification rule:

\begin{align}\label{eq:ChainDecisionRuleSimpleBR}
h^{BR}_{\pi(i)}(\vec{x}) &=\argmax{y\in \{0,1\}}P(\textbf{Y}_{\pi(i)}=y)\prod_{m=1}^{d}P(\vec{\textbf{X}}_{m}=\vec{x}_m|\textbf{Y}_{\pi(i)}=y)
\end{align}

Since the neighbourhood of a given instance is defined as a fuzzy set, consistently the above-mentioned sets are also defined as fuzzy. However, the sets are fuzzy singletons. The visualisation of aforementioned sets is provided in \figurename~\ref{figure:Venn}.
{
\begin{figure}[tb]
\begin{center}
   \includegraphics[width=0.2\textwidth]{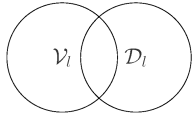}
   \caption{Visualisation of ground truth labels and the decision set of the algorithm.}%
   \label{figure:Venn}
\end{center}
\end{figure}
}

Using the above-mentioned sets we define local True Positive rate, False Positive rate, False Negative rate respectively:
\begin{align}
  \mathrm{TP}_{l}(\vec{x}) &= \left| \mathcal{V}_l \cap \mathcal{D}_{l} \cap \mathcal{N}(\vec{x}) \right|, \\
  \mathrm{FP}_{l}(\vec{x}) &= \left| \left( \mathcal{D}_{l}\setminus \mathcal{V}_{l} \right) \cap \mathcal{N}(\vec{x}) \right|,\\
  \mathrm{FN}_{l}(\vec{x}) &= \left| \left(  \mathcal{V}_{l} \setminus \mathcal{D}_{l}  \right) \cap \mathcal{N}(\vec{x}) \right|,
 \end{align}
where $|\cdot|$ is the cardinality of a fuzzy set~\cite{Dhar2013}.
Then, we define the local measure of classification quality:
\begin{equation}\label{eq:F_local}
 F_{l}(\vec{x}) = \dfrac{2\mathrm{TP}_{l}(\vec{x})}{ 2\mathrm{TP}_{l}(\vec{x}) + \mathrm{FP}_{l}(\vec{x}) + \mathrm{FN}_{l}(\vec{x}) }
\end{equation}

Finally, the label order $\pi$ is chosen so that the following inequalities are met:
\begin{equation}
 F_{\pi(1)}(\vec{x})\geq F_{\pi(2)}(\vec{x}) \geq \cdots \geq F_{\pi(L)}(\vec{x}).
\end{equation}
That is labels for whom the classification quality is higher precedes other labels in the chain structure. In other words, this simple heuristic is aimed at dealing with error propagation in the chain structure by employing the most accurate models at the beginning of the chain. A similar approach based on the Jaccard quality criterion was proposed in~\cite{Soonsiripanichkul2016}.

\subsection{The Ensemble Classifier}\label{subsec:EnsClassifier}

Now, let us define a ML $K-\mathrm{element}$ classifier ensemble: $eH = \left\{ H_1,\dots, H_K \right\}$. The ensemble is built using classifier chain algorithms defined in previous sections. Each ensemble classifier is built using a subset of the original dataset. The size of subset is $66\%$ of the original training set. 

The BR transformation may produce imbalanced single-label dataset. To prevent the classifier from learning from a highly imbalanced dataset, we applied the random undersampling technique~\cite{Garca2012}. The majority class is undersampled when imbalance ratio is higher than 20. The goal of undersampling is to keep the imbalance ratio at the level of 20.   

The research on the application of Naive Bayes algorithm under the CC framework shows that when the number of features in the input space is significantly higher in comparison to the number of labels the Naive Bayes classifier may not perform well~\cite{daSilva2014}. To prevent the proposed system from being affected by this phenomenon, we applied the feature selection procedure for each single-label separately. That is, the attributes are selected in order to improve the classification quality for given label. The feature selection removes only attributes related to the original input space. Features related to labels are passed through the chain without selection. We employed the selection procedure based on correlation. In other words, we select attributes that are highly correlated to the predicted label and their inter-correlations are low~\cite{Hall1999}. Additionally, if the number of selected features is higher than 300, we select 300 random features from the set of previously selected features. 

The final prediction vector of the ensemble is obtained via is a simple averaging of response vectors corresponding to base classifiers of the ensemble followed by the thresholding procedure: 
\begin{equation}
	\widetilde{h}_{i}(\vec{x}) = \ivBrack{K^{-1}\sum_{k=1}^{K}h_{i}^{k}(\vec{x}) >0.5},
\end{equation}
where $\ivBrack{\cdot}$ is the Iverson bracket.

\section{Experimental Setup}\label{sec:ExpSetup}

The experimental study is divided into three main sections. The first one assesses the impact of employing chaining approach. In the section, we compare binary relevance and classifier chains algorithms built using the following base classifiers:
\begin{itemize}
 \item J48 Classifier (C4.5 algorithm implemented in Weka)~\cite{Quinlan1993}.
 \item SVM algorithm with radial based kernel~\cite{Cortes1995,CC01a}.
 \item Naive Bayes Classifier~\cite{Hand2001}.
 \item Nearest Neighbour classifier~\cite{Cover1967}.
\end{itemize}
In this section, we compare BR and CC ensembles built using a genetic algorithm tailored to optimise the macro-averaged $F_1$ loss. For each ensemble, the size of the committee was set to $K=20$. For the algorithm based on the genetic algorithm, the initial size of the committee was set to $3K$. Each numeric attribute in the training and validation datasets was also standardised. After the standardisation, the mean value of the attribute is 0 and its standard deviation is 1.

During the experimental study, the parameters of the SVM classifier ($C \in \{0.001,1,2,\cdots, 10 \}$, $\gamma \in \{0.001,1,2,\cdots, 5\}$) were tuned using grid search and 3-fold cross validation. The number of nearest neighbours was also tuned using 3-fold cross validation. The number of neighbours was chosen among the following values $R \in \{1,3,5,\cdots, 11 \}$.

In two remaining sections, the conducted experimental study provides an empirical evaluation of the classification quality of the proposed methods and compares it to reference methods. Namely, we conducted our experiments using the following algorithms of building a CC ensemble:
\begin{enumerate}
 \item The proposed approach (Section~\ref{subsec:DynamicCC}).
 \item Static ensemble generated using a genetic algorithm~\cite{Trajdos2017}. The enesmble is tuned to optimise the macro-averaged $F_1$ measure
 \item ECC ensemble with randomly generated chain orders~\cite{Read2009}.
 \item OOCC dynamic method proposed by da Silva et al.~\cite{daSilva2014}. The ensemble is tuned to optimise the example based $F_1$ measure. Additionally, the reference method uses single split into training and validation sets. 
\end{enumerate}
The above-mentioned methods of building CC systems were evaluated using Naive Bayes and the nearest neighbour algorithms as base classifiers. Systems built using different base classifiers are investigated in two separate sections. In the sections, we will refer to the investigated algorithms using the above-said numbers.

The reference algorithm also uses Naive Bayes/nearest neighbour algorithm with data preprocessing procedures described in Section~\ref{subsec:EnsClassifier}. 

The extraction of training and test datasets was performed using $10$ fold cross-validation. For each ensemble, the proportion of the training set $\mathcal{T}_A$ was fixed at $t=0.6$ of the original training set (see Algorithm~\ref{fig:TrainNB}). For each ensemble, the size of the committee was set to $K=20$. For the algorithm based on the genetic algorithm, the initial size of the committee was set to $3K$. Each numeric attribute in the training and validation datasets was also standardised. After the standardisation, the mean value of the attribute is 0 and its standard deviation is 1.

The $\beta$ coefficient was tuned during the training procedure using 3 CV approach. The best value among $\{1,2,\cdots, 10 \}$ is chosen. 

Single label classifiers were implemented using WEKA software~\cite{Hall2009}. Multi-label classifier were implemented using Mulan software~\cite{mulan_software}.

The experiments were conducted using 30 multi-label benchmark sets. The main characteristics of the datasets are summarized in Table~\ref{table:Dataset_summ}. We used datasets from the sources abbreviated as follows:A~\cite{Charte2015}, B~\cite{meka} M--\cite{Tsoumakas2011_mulan}; W--\cite{Wu2014}; X--\cite{Xu2013}; Z--\cite{Zhou2012}; T--\cite{Tomas2014}; O -- \cite{thorsten1998}. Some of the employed sets needed some preprocessing. That is, we used multi-label multi-instance~\cite{Zhou2012} sets (sources Z and W) which were transformed to single-instance multi-label datasets according to the suggestion made by Zhou et al.~\cite{Zhou2012}. Multi-target regression sets (No 9, 30) were binarised using simple thresholding strategy. That is if the response is greater than 0 the resulting label is set relevant. Two of the used datasets are synthetic ones (source T) and they were generated using algorithm described in~\cite{Tomas2014}. To reduce the computational burden, we use only a subset of original Tmc2007 and IMDB sets. Additionally, the number of labels in Stackex datasets is reduced to 15. 

The algorithms were compared in terms of 11 different quality criteria coming from three groups~\cite{Luaces2012}: Instance-based (Hamming, Zero-One, $F_1$, False Discovery Rate, False Negative Rate); Label-based. The last group contains the following measures:  Macro Averaged (False Discovery Rate (FDR, 1- Precision), False Negative Rate (FNR, 1-Recall), $F_1$) and Micro Averaged versions of the above-mentioned criteria.

Statistical evaluation of the results was performed using the Wilcoxon signed-rank test~\cite{demsar2006,wilcoxon1945} and the family-wise error rates were controlled using the Holm procedure~\cite{demsar2006,holm1979}. For all statistical tests, the significance level was set to $\alpha=0.1$. Additionally, we also applied the Friedman~\cite{Friedman1940} test followed by the Nemenyi post-hoc procedure~\cite{demsar2006}.
{
\def\arraystretch{0.5}
\begin{table}
 \centering\scriptsize
 \caption{Summarised properties of the datasets employed in the experimental study. Sr denotes the source of dataset, No. is the ordinal number of a set, $N$ is the number of instances, $d$ is the dimensionality of input space, $L$ denotes the number of labels. $\mathrm{LC}$, $\mathrm{LD}$, $\mathrm{avIR}$ are label cardinality, label density and  average imbalance ratio respectively~\cite{Luaces2012,Charte2014}.}
 \label{table:Dataset_summ}
 \begin{tabular}{lllllllll}
No&Name&Sr&$N$&$d$&$L$&LC&LD&avIR\\
\toprule
1&Arts1&	M	&7484&1733&26&1.654&.064&94.74\\
2&Azotobacter&	W	&407&20&13&1.469&.113&2.225\\
3&Birds&	M	&645&260&19&1.014&.053&5.407\\
4&Caenorhabditis&	W	&2512&20&21&2.419&.115&2.347\\
5&Drosophila&	W	&2605&20&22&2.656&.121&1.744\\
6&Emotions&	M	&593&72&6&1.868&.311&1.478\\
7&Enron&	M	&1702&1001&53&3.378&.064&73.95\\
8&Flags&	X	&194&43&7&3.392&.485&2.255\\
9&Flare2&	M	&1066&27&3&0.209&.070&14.15\\
10&Genbase&	M	&662&1186&27&1.252&.046&37.32\\
11&Geobacter&	W	&379&20&11&1.264&.115&2.750\\
12&Haloarcula&	W	&304&20&13&1.602&.123&2.419\\
13&Human&	X	&3106&440&14&1.185&.085&15.29\\
14&Image&	M	&2000&294&5&1.236&.247&1.193\\
15&IMDB&	M	&3042&1001&28&1.987&.071&24.61\\
16&LLOG&	B	&1460&1004&75&1.180&.016&39.27\\
17&Medical&	M	&978&1449&45&1.245&.028&89.50\\
18&MimlImg&	Z	&2000&135&5&1.236&.247&1.193\\
19&Ohsumed&	O	&13929&1002&23&1.663&.072&7.869\\
20&Plant&	X	&978&440&12&1.079&.090&6.690\\
21&Pyrococcus&	W	&425&20&18&2.136&.119&2.421\\
22&Saccharomyces&	W	&3509&20&27&2.275&.084&2.077\\
23&Scene&	X	&2407&294&6&1.074&.179&1.254\\
24&SimpleHC&	T	&3000&30&10&1.900&.190&1.138\\
25&SimpleHS&	T	&3000&30&10&2.307&.231&2.622\\
26&SLASHDOT&	B	&3782&1079&22&1.181&.054&17.69\\
27&Stackex\_chess&	A	&1675&585&15&1.137&.076&4.744\\
28&Tmc2007-500&	M	&2857&500&22&2.222&.101&17.15\\
29&water-quality&	M	&1060&16&14&5.073&.362&1.767\\
30&yeast&	M	&2417&103&14&4.237&.303&7.197\\
 \bottomrule
\end{tabular}
 
\end{table}
}

\section{Results and Discussion}\label{sec:ResDisc}

\subsection{Assessing the impact of chaining approach}\label{sec:ResDisc:ChainAppImpact}

In this section, we assess the consequences of employing chaining approach. That is, we compare binary relevance ensembles with classifier chain ensembles built using the same base classifier. The results are shown in Figure~\ref{figure:critRadBaseCompare} and Table~\ref{table:StatEval_Base}. Full results are presented in the appendix in Tables~\ref{table:FullResults1Base}, \ref{table:FullResults2Base} and~\ref{table:FullResults3Base}.
The compared algorithms are numbered as follows:
\begin{enumerate}
 \item BR ensemble built using J48 algorithm.
 \item CC ensemble built using J48 algorithm.
 \item BR ensemble built using SVM algorithm.
 \item CC ensemble built using SVM algorithm.
 \item BR ensemble built using NB algorithm.
 \item CC ensemble built using NB algorithm.
 \item BR ensemble built using KNN algorithm.
 \item CC ensemble built using KNN algorithm.
\end{enumerate}
The analysis of the results clearly shows that there is a noticeable difference between two groups of measures. That is, the differences between BR-based and CC-based algorithms are greater in terms of example based criteria. On the other hand, the differences in mean ranks are lower for example based measures. 

For the example based measures, the average ranks achieved by CC-based algorithms are lower than for BR-based algorithms. However, only for algorithms based on J48 classifier, the differences are significant for example-based FDR, FNR and $F_1$ measures.  A similar trend is observed for the zero-one loss. In this case, only differences for the nearest neighbour classifier are insignificant. It means that CC-based classifiers obtain the higher number of 'perfect match' results. 

On the other hand, for the label-based measures and Hamming loss, almost no significant differences are observed. However, the average ranks suggest that for this group of measures, the classification quality may deteriorate. 

The results clearly show that although label-specific quality measures do not change in a significant way, the prediction of the entire label-vector improves. This is an expected result since the CC-based approach incorporates the inter-label relations. This is a well-known fact that has been reported by authors that have previously compared both approaches~\cite{Madjarov2012}.

The results also show that there are almost no significant differences between J48, NB and KNN based algorithms. Contrary, SVM algorithm tends to outperform the remaining ones in terms of example-based criteria, hamming loss and zero-one loss. It means that although J48 algorithm takes the biggest advantage of employing the chain rule,  NB and KNN based classifiers are comparable to J48-based ensembles.

{
\begin{figure}[tb]
\begin{center}
   \includegraphics[width=0.5\textwidth]{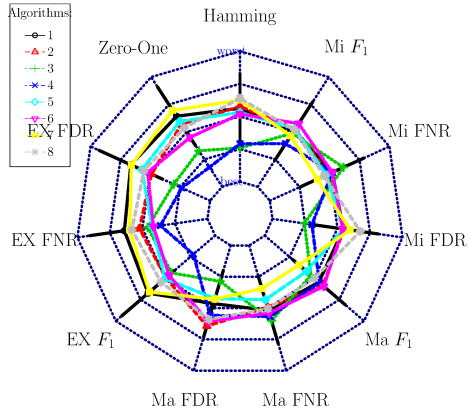}
   \caption{Base classifier comparison. Visualisation of average ranks achieved by algorithms and corresponding critical distances for the Nemenyi post-hoc test. Each axis of the radar plot corresponds to given quality criterion. The closer a point is to the centre of the radar plot, the lower average rank is (lower is better). Black bars parallel to criterion-specific axes denote critical difference for the Nemenyi tests.}%
   \label{figure:critRadBaseCompare}
\end{center}
\end{figure}
}

{
\setlength\tabcolsep{0.5pt}
\def\arraystretch{0.8}
\begin{sidewaystable}
\centering\scriptsize
\caption{Base classifier comparison. Result of the statistical evaluation. Rnk stands for average rank over all datasets, Frd is the p-value obtained using the Friedman test and Wp-i denotes the p-value associated with the Wilcoxon test that compares the i-th algorithm against the others.}\label{table:StatEval_Base}
\begin{tabular}{c|cccccccc|cccccccc|cccccccc|cccccccc}
  Alg.&1&2&3&4&5&6&7&8	&1&2&3&4&5&6&7&8	&1&2&3&4&5&6&7&8	&1&2&3&4&5&6&7&8\\
  \hline
   &\multicolumn{8}{c|}{Hamming} & \multicolumn{8}{c|}{Zero-One}& \multicolumn{8}{c|}{EX FDR} & \multicolumn{8}{c}{EX FNR}\\ 
  \hline
   Rnk &4.94&5.03&2.81&3.03&4.72&4.63&5.38&5.47	&5.50&5.00&3.30&2.25&5.17&4.16&5.88&4.75	&5.64&4.44&3.13&2.66&5.00&4.66&5.61&4.88		&5.47&4.66&4.22&3.53&3.97&3.91&5.19&5.06\\
   Frd&\multicolumn{8}{c|}{$.000018$ }&	\multicolumn{8}{c|}{$.000000$}	&\multicolumn{8}{c|}{$.000002$}	&\multicolumn{8}{c}{$.056423$}\\ 
   Wp-1&  & 1.00 & .002 & .148 & 1.00 & 1.00 & 1.00 & .432	&  & .032 & .001 & .000 & 1.00 & .478 & 1.00 & 1.00 	&  & .070 & .001 & .001 & 1.00 & .650 & 1.00 & 1.00	&  & .011 & 1.00 & .009 & .205 & .054 & 1.00 & 1.00	\\
   Wp-2&  &  & .000 & .025 & 1.00 & 1.00 & 1.00 & 1.00	&  &  & 1.00 & .001 & 1.00 & 1.00 & .044 & 1.00	&  &  & .845 & .001 & 1.00 & 1.00 & .056 & 1.00	&  &  & 1.00 & .223 & 1.00 & 1.00 & 1.00 & 1.00\\
   Wp-3 &  &  &  & 1.00 & .014 & .022 & .000 & .000	&  &  &  & .032 & .005 & .627 & .001 & .597 	&  &  &  & 1.00 & .002 & .097 & .001 & .573 	&  &  &  & .937 & 1.00 & 1.00 & 1.00 & 1.00\\
   Wp-4&  &  &  &  & .552 & .552 & .017 & .002	&  &  &  &  & .000 & .004 & .000 & .002	&  &  &  &  & .002 & .006 & .000 & .002	&  &  &  &  & 1.00 & 1.00 & .169 & .113\\
   Wp-5&  &  &  &  &  & 1.00 & 1.00 & .591	&  &  &  &  &  & .041 & 1.00 & 1.00	&  &  &  &  &  & 1.00 & .964 & 1.00	&  &  &  &  &  & 1.00 & .221 & 1.00\\
   Wp-6&  &  &  &  &  &  & 1.00 & 1.00 	&  &  &  &  &  &  & .478 & 1.00	&  &  &  &  &  &  & .666 & 1.00	&  &  &  &  &  &  & .090 & .266 \\
   Wp-7&  &  &  &  &  &  &  & 1.00 	&  &  &  &  &  &  &  & .220	&  &  &  &  &  &  &  & 1.00	&  &  &  &  &  &  &  & 1.00\\
 \hline
 & \multicolumn{8}{c|}{EX $F_1$}& \multicolumn{8}{c|}{Macro FDR} &\multicolumn{8}{c|}{Macro FNR}&\multicolumn{8}{c}{Macro $F_1$}\\
 \hline
 Rnk &5.72&4.47&4.06&2.53&4.53&4.25&5.63&4.81	&4.25&5.50&3.00&4.91&3.97&5.25&4.00&5.13	&4.63&4.45&5.20&4.97&4.03&4.78&3.41&4.53		&4.81&5.09&4.25&4.84&3.94&5.16&3.41&4.50\\
   Frd&\multicolumn{8}{c|}{$0.000030$ }&	\multicolumn{8}{c|}{$0.002491$}	&\multicolumn{8}{c|}{$0.312535$}	&\multicolumn{8}{c}{$0.224854$}\\ 
   Wp-1	&  & .005 & .041 & .000 & .200 & .072 & 1.00 & 1.00	&  & .019 & .552 & 1.00 & 1.00 & .390 & 1.00 & .479	&  & 1.00 & 1.00 & 1.00 & 1.00 & 1.00 & .022 & 1.00	&  & 1.00 & 1.00 & 1.00 & 1.00 & 1.00 & .000 & 1.00 \\
   Wp-2	&  &  & 1.00 & .001 & 1.00 & 1.00 & .189 & 1.00	&  &  & .003 & 1.00 & .697 & 1.00 & .152 & 1.00	&  &  & 1.00 & 1.00 & 1.00 & 1.00 & .441 & 1.00	&  &  & 1.00 & 1.00 & 1.00 & 1.00 & .031 & 1.00 \\
   Wp-3	&  &  &  & .272 & 1.00 & 1.00 & .089 & 1.00	&  &  &  & .066 & .552 & .023 & .349 & .003	&  &  &  & 1.00 & .077 & 1.00 & .058 & 1.00	&  &  &  & 1.00 & 1.00 & 1.00 & 1.00 & 1.00 \\
   Wp-4	&  &  &  &  & .014 & .007 & .000 & .005	&  &  &  &  & 1.00 & 1.00 & 1.00 & 1.00	&  &  &  &  & .419 & 1.00 & .501 & 1.00	&  &  &  &  & 1.00 & 1.00 & .824 & 1.00 \\
   Wp-5	&  &  &  &  &  & 1.00 & .200 & 1.00	&  &  &  &  &  & .208 & 1.00 & 1.00	&  &  &  &  &  & .998 & 1.00 & 1.00	&  &  &  &  &  & .355 & 1.00 & 1.00 \\
   Wp-6	&  &  &  &  &  &  & .090 & 1.00	&  &  &  &  &  &  & .101 & 1.00	&  &  &  &  &  &  & 1.00 & 1.00	&  &  &  &  &  &  & .278 & 1.00 \\
   Wp-7	&  &  &  &  &  &  &  & 1.00	&  &  &  &  &  &  &  & .552	&  &  &  &  &  &  &  & .143	&  &  &  &  &  &  &  & .203 \\
   \hline
     &\multicolumn{8}{c|}{Micro FDR}&\multicolumn{8}{c|}{Micro FNR}&\multicolumn{8}{c|}{Micro $F_1$}\\
     \hhline{-------------------------}
  Rnk	&4.81&4.84&2.72&3.13&4.81&4.78&5.19&5.72	&4.66&4.44&5.31&4.66&4.31&4.66&3.78&4.19		&4.56&4.69&4.41&3.75&4.69&5.03&4.25&4.63\\
  Frd&\multicolumn{8}{c|}{$0.000014$ }&	\multicolumn{8}{c|}{$0.777721$}	&\multicolumn{8}{c|}{$0.777721$}\\ 
	Wp-1&  & 1.00 & .001 & .141 & 1.00 & 1.00 & 1.00 & .045	&  & 1.00 & .706 & 1.00 & 1.00 & 1.00 & .136 & 1.00	&  & 1.00 & 1.00 & 1.00 & 1.00 & 1.00 & 1.00 & 1.00 &  &  &  &  &  &  &  &  \\
	Wp-2&  &  & .000 & .023 & 1.00 & 1.00 & 1.00 & .564	&  &  & 1.00 & 1.00 & 1.00 & 1.00 & 1.00 & 1.00	&  &  & 1.00 & 1.00 & 1.00 & 1.00 & 1.00 & 1.00 &  &  &  &  &  &  &  &  \\
	Wp-3&  &  &  & 1.00 & .001 & .001 & .000 & .001	&  &  &  & 1.00 & .233 & 1.00 & .215 & .132	&  &  &  & 1.00 & 1.00 & 1.00 & 1.00 & 1.00 &  &  &  &  &  &  &  &  \\
	Wp-4&  &  &  &  & .091 & .079 & .015 & .007	&  &  &  &  & 1.00 & 1.00 & 1.00 & 1.00	&  &  &  &  & 1.00 & 1.00 & 1.00 & 1.00 &  &  &  &  &  &  &  &  \\
	Wp-5&  &  &  &  &  & 1.00 & 1.00 & 1.00	&  &  &  &  &  & 1.00 & 1.00 & 1.00	&  &  &  &  &  & 1.00 & 1.00 & 1.00 &  &  &  &  &  &  &  &  \\
	Wp-6&  &  &  &  &  &  & 1.00 & 1.00	&  &  &  &  &  &  & 1.00 & 1.00	&  &  &  &  &  &  & 1.00 & 1.00 &  &  &  &  &  &  &  &  \\
	Wp-7&  &  &  &  &  &  &  & 1.00	&  &  &  &  &  &  &  & 1.00	&  &  &  &  &  &  &  & 1.00 &  &  &  &  &  &  &  &  \\
  \hhline{-------------------------}
   
\end{tabular}
\end{sidewaystable}
}

\subsection{Naive Bayes Classifier}\label{sec:ResDisc:NB}
\begin{figure}[tb]
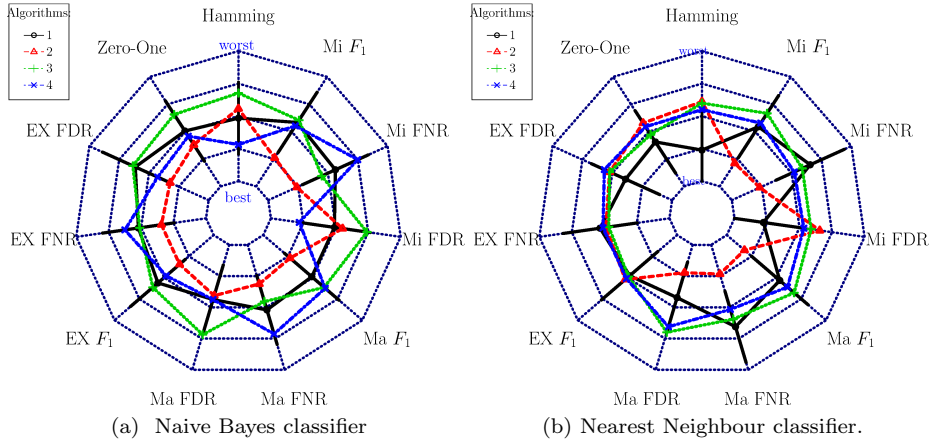

\begin{center}
\subfloat[ Naive Bayes classifier]{\label{figure:critRad}%
  \includegraphics[width=0.5\textwidth]{\ptFiguresDirectory{radar}}%
}
\subfloat[Nearest Neighbour classifier.]{\label{figure:critRadKNN}%
  \includegraphics[width=0.5\textwidth]{\ptFiguresDirectory{radar_KNN}}%
}

\caption{Visualisation of average ranks achieved by algorithms.} 
\end{center}
\end{figure}

The results of the experimental study are presented in Table~\ref{table:StatEval} and Figure~\ref{figure:critRad}. Tables~\ref{table:FullResults1},~\ref{table:FullResults2Base} and~\ref{table:FullResults3} show full results of the experiment. Table~\ref{table:StatEval} provides results of the statistical evaluation of the experiments. Figure~\ref{figure:critRad} visualises the average ranks and provide a view of the Nemenyi post-hoc procedure. 

First, let's analyse differences between the proposed heuristic and the simple ECC ensemble. The proposed method is tailored to optimise the macro-averaged $F_1$ loss so we begin with investigating macro-averaged measures. It is easy to see that both methods are comparable in terms of recall but the proposed one is significantly better in terms of precision. It means that the proposed method makes significantly less false positive predictions. Consequently, under the macro-averaged $F_1$ loss the proposed method outperforms the ECC ensemble. The same pattern is also present in results related to micro-averaged measures. However, the difference for the micro-averaged $F_1$ measure is not significant. 
In contrast, under example based measures, except the Hamming loss,  there are no significant differences between investigated methods. 

The results show that the proposed heuristic provides an effective way of improving classification quality for classifier chains ensemble. Moving the best performing label-specific models at the beginning of the chain reduces the error that propagates along the chain. What is more, the experimental study also showed that the Naive Bayes classifier combined with proper data preprocessing may be effectively employed in classifier chain ensembles.

Now, let's compare the proposed method to the other algorithm based on the dynamic chain approach. When we investigate the example-based criteria it is easy to see that the OOCC algorithm outperforms the proposed one in terms of FDR and Hamming loss. Those results combined with results achieved in terms of macro and micro averaged measures shows that the OOCC mthod seems to be too much conservative.  That is, it tends to makes many false negative predictions in comparison to the other methods. The outstanding results for the Hamming loss are a consequence of the imbalanced nature of the multi-label data.  That is, the presence of labels is relatively rare and the prediction that contains many false negatives may achieve inadequately hight performance under the Hamming loss~\cite{Luaces2012}.

On the other hand, the average ranks clearly show that the method based on genetic algorithm achieves the best results in comparison to the other investigated methods. The main reason is that the GA-based approach optimises the entire ensemble structure, whereas the investigated dynamic chain methods, choose the best label order for single classifier chain. Then the locally chosen chains are combined into an ensemble. It gives us an important clue.  That is when we consider an algorithm for dynamic chain order selection, we should think about a single chain and the global structure of the entire ensemble as well. 

{
\setlength\tabcolsep{1.9pt}
\def\arraystretch{0.8}
\begin{table}
\centering\scriptsize
\caption{Naive Bayes Based CC. Results of the statistical evaluation.}\label{table:StatEval}
\begin{tabular}{c|cccc|cccc|cccc|cccc}
  Alg.&1&2&3&4	&1&2&3&4	&1&2&3&4	&1&2&3&4\\
  \hline
   &\multicolumn{4}{c|}{Hamming} & \multicolumn{4}{c|}{Zero-One}& \multicolumn{4}{c|}{EX FDR} & \multicolumn{4}{c}{EX FNR}\\ 
  \hline
  Rnk& 2.45 & 2.67 & 3.03 & \textBF{1.84}	& 2.52 & \textBF{2.14} & 2.97 & 2.38 	& 2.84 & \textBF{1.97} & 2.91 & 2.28	& 2.56 & \textBF{2.03} & 2.53 & 2.88 \\
  Frd&\multicolumn{4}{c|}{$.02113$ }&	\multicolumn{4}{c|}{$.19410$}	&\multicolumn{4}{c|}{$.03824$}	&\multicolumn{4}{c}{$.19410$}\\ 
  Wp-1&  & .702 & .025 & .063		&  & .358 & .199 & .761	&  & .019 & .295 & .134 &  & .053 & .700 & .700 \\
  Wp-2 &  &  & .319 & .319	&  &  & .006 & .368	&  &  & .005 & .239	&  &  & .136 & .015\\
  Wp-3 &  &  &  & .001	&  &  &  & .097	&  &  &  & .040	&  &  &  & .136\\
  \hline
   & \multicolumn{4}{c|}{EX $F_1$}& \multicolumn{4}{c|}{Macro FDR} &\multicolumn{4}{c|}{Macro FNR}&\multicolumn{4}{c}{Macro $F_1$}\\
  \hline
  Rnk& 2.69 & \textBF{2.03} & 2.84 & 2.44	& 2.31 & \textBF{2.22} & 3.16 & 2.31 	& 2.56 & \textBF{1.94} & 2.36 & 3.14	& 2.47 & \textBF{1.81} & 2.84 & 2.88 \\
  Frd&\multicolumn{4}{c|}{$.19410$ }&	\multicolumn{4}{c|}{$.04402$}	&\multicolumn{4}{c|}{$.02113$}	&\multicolumn{4}{c}{$.02113$}\\ 
  Wp-1&  & .066 & .821 & .821		&  & 1.00 & .012 & 1.00	&  & .035 & .919 & .022	&  & .156 & .096 & .248 \\
  Wp-2&  &  & .017 & .112		&  &  & .031 & 1.00		&  &  & .174 & .002		&  &  & .003 & .022 \\
  Wp-3&  &  &  & .590			&  &  &  & .012		&  &  &  & .105		&  &  &  & .733 \\
  \hline
   &\multicolumn{4}{c|}{Micro FDR}&\multicolumn{4}{c|}{Micro FNR}&\multicolumn{4}{c|}{Micro $F_1$}\\
  \hhline{-------------}
  Rnk& 2.47 & 2.66 & 3.22 & \textBF{1.66}	& 2.69 & \textBF{1.72} & 2.34 & 3.25 	& 2.75 & \textBF{1.78} & 2.81 & 2.66\\
  Frd&\multicolumn{4}{c|}{$.00027$ }&	\multicolumn{4}{c|}{$.00029$}	&\multicolumn{4}{c|}{$.02242$}\\ 
  Wp-1&  & 1.00 & .028 & .000		&  & .005 & .254 & .239	&  & .044 & .610 & 1.00\\
  Wp-2&  &  & 1.00 & .008		&  &  & .052 & .000		&  &  & .002 & .044\\
  Wp-3&  &  &  & .000			&  &  &  & .014		&  &  &  & 1.00\\
  \hhline{-------------}

\end{tabular}
\end{table}
}

\subsection{Nearest Neighbour Classifier}\label{sec:ResDisc:KNN}
The results of the experimental study are presented in Table~\ref{table:StatEvalKNN} and Figure~\ref{figure:critRadKNN}. Tables~\ref{table:FullResults1KNN},~\ref{table:FullResults2KNN} and~\ref{table:FullResults3KNN} show full results of the experiment. Table~\ref{table:StatEval} provides results of the statistical evaluation of the experiments. Figure~\ref{figure:critRadKNN} visualises the average ranks and provide a view of the Nemenyi post-hoc procedure. 

The results show that for the group of example based measures and the zero-one loss, there are no significant differences in classification quality between all investigated algorithms. 

For macro and micro averaged measures, the best performing algorithm is an ensemble optimised using the genetic algorithm. The proposed nearest-neighbour-based classifier does not differ significantly from ECC OOCC algorithms. However, it tends to be more conservative because it achieves lower FDR and higher FDR. In other words, the classifier tends to decrease the false positive rate at the cost of decreasing the true positive rate. This phenomenon causes the highest classification quality in terms of the Hamming loss. The reason is that for the multi-label set with low label density, it is easy to obtain high classification by setting all possible labels as irrelevant. 

The results confirm the findings described in Section~\ref{sec:ResDisc:ChainAppImpact}. The nearest-neighbour-based CC algorithm is unable to take all benefits of the chaining approach. On the other hand, the method is still comparable to chains built using different base classifiers. The main goal of this paper was to propose the model that can change the chain structure without retraining. This goal was achieved. 

{
\setlength\tabcolsep{1.9pt}
\begin{table}
\centering\scriptsize
\caption{Nearest Neighbour based CC. Result of statistical evaluation.}\label{table:StatEvalKNN}
\begin{tabular}{c|cccc|cccc|cccc|cccc}
  Alg.&1&2&3&4	&1&2&3&4	&1&2&3&4	&1&2&3&4\\
  \hline
   &\multicolumn{4}{c|}{Hamming} & \multicolumn{4}{c|}{Zero-One}& \multicolumn{4}{c|}{EX FDR} & \multicolumn{4}{c}{EX FNR}\\ 
  \hline
  Rnk& 1.72 & 2.83 & 2.80 & 2.65	& 2.22 & 2.73 & 2.42 & 2.63 	& 2.17 & 2.57 & 2.57 & 2.70 	& 2.60 & 2.47 & 2.40 & 2.53 \\
  Frd&\multicolumn{4}{c|}{$.008583$ }&	\multicolumn{4}{c|}{$1.00000$}	&\multicolumn{4}{c|}{$1.00000$}	&\multicolumn{4}{c}{$1.00000$}\\ 
  Wp-1&  & .004 & .000 & .006 &  & 1.00 & 1.00 & 1.00	&  & 1.00 & 1.00 & .948	&  & 1.00 & .985 & 1.00 \\
  Wp-2 &  &  & .441 & .173  &  &  & 1.00 & 1.00	&  &  & 1.00 & 1.00	&  &  & 1.00 & 1.00 \\
  Wp-3&  &  &  & .441  &  &  &  & 1.00	&  &  &  & 1.00	&  &  &  & 1.00 \\
  \hline
   & \multicolumn{4}{c|}{EX $F_1$}& \multicolumn{4}{c|}{Macro FDR} &\multicolumn{4}{c|}{Macro FNR}&\multicolumn{4}{c}{Macro $F_1$}\\
  \hline
  Rnk& 2.43 & 2.57 & 2.47 & 2.53	& 2.27 & 1.67 & 3.10 & 2.97 	& 2.97 & 1.70 & 2.78 & 2.55	
 & 2.600 & 1.533 & 3.033 & 2.833 \\
  Frd&\multicolumn{4}{c|}{$1.00000$ }&	\multicolumn{4}{c|}{$.000265$}	&\multicolumn{4}{c|}{$.004979$}	&\multicolumn{4}{c}{$.000259$}\\ 
  Wp-1	&  & 1.00 & 1.00 & 1.00 	&  & .017 & .010 & .012	&  & .000 & .617 & .617	&  & .000 & .276 & .579 \\
  Wp-2	&  &  & 1.00 & 1.00 	&  &  & .000 & .003	&  &  & .000 & .004	&  &  & .000 & .000 \\
  Wp-3	&  &  &  & 1.00  	&  &  &  & .271	&  &  &  & .561	&  &  &  & .579 \\
  \hline
   &\multicolumn{4}{c|}{Micro FDR}&\multicolumn{4}{c|}{Micro FNR}&\multicolumn{4}{c|}{Micro $F_1$}\\
  \hhline{-------------}
  Rnk& 1.67 & 2.97 & 2.77 & 2.60	& 2.97 & 1.70 & 2.77 & 2.57 	& 2.63 & 1.63 & 3.00 & 2.73\\
  Frd&\multicolumn{4}{c|}{$ .003725$ }&	\multicolumn{4}{c|}{$.004979$}	&\multicolumn{4}{c|}{$.002113$}\\ 
  Wp-1	&  & .009 & .000 & .028 	&  & .000 & .926 & .926	&  & .002 & .532 & .657 \\
  Wp-2	&  &  & .382 & .070  	&  &  & .002 & .019	&  &  & .001 & .006 \\
  Wp-3	&  &  &  & .715  	&  &  &  & .926	&  &  &  & .657 \\
  \hhline{-------------}

\end{tabular}
\end{table}
}

\section{Conclusions}\label{sec:Conclusions}

The main goal of this research was to provide an effective chain classifier that allows changing label order at relatively low computational cost. We achieved it using a classifier based on the Naive Bayes approach. To prove that the proposed method allows handling inter-label relations in an efficient way, we proposed a simple heuristic method that determines label order that should minimise label propagation error. Indeed, the experimental results showed that the proposed method is able to produce a good chain structure at a low computational cost. However, the proposed method of building a dynamic ensemble does not allow to outperform the static system that optimizes the entire ensemble structure. 
The obtained results are very promising. We believe that there is still a room for improvement.  In our opinion, the performance of the system may be improved if we provide better, a better heuristic that optimises the entire ensemble in a dynamic way.  The proposed dynamic classifier is a first step in the process of investigating dynamic classifier chain ensembles.

Another way of improving this idea is to build different classifiers that would be able to change the chain structure without retraining the entire model.

\section*{Acknowledgment}

The work was supported by the statutory funds of the Department of Systems and Computer Networks, Wroclaw University of Science and Technology, under agreement 0401/0159/16.

\bibliography{bibliography}
\appendix
\section{Full results}\label{sec:app:BaseClassCopar}

{
\setlength\tabcolsep{1.5pt}
\def\arraystretch{0.8}
\begin{sidewaystable}
\centering\scriptsize
\caption{Base classifier comparison -- full results -- example based criteria.}\label{table:FullResults1Base}
\begin{tabular}{c|cccccccc|cccccccc|cccccccc}
&\multicolumn{8}{c|}{Macro FDR}&\multicolumn{8}{c|}{Macro FNR}&\multicolumn{8}{c}{Macro $F_1$}\\
No.&1&2&3&4&5&6&7&8	&1&2&3&4&5&6&7&8	&1&2&3&4&5&6&7&8\\
\hline

1&.668&.622&.758&.645&.789&.780&.666&.590&.682&.642&.775&.604&.802&.782&.692&.622&.694&.651&.781&.649&.810&.796&.697&.625\\
2&.639&.608&.507&.527&.568&.565&.722&.664&.638&.593&.529&.530&.564&.561&.678&.623&.663&.620&.528&.534&.575&.570&.729&.678\\
3&.386&.402&.384&.418&.367&.389&.394&.402&.408&.429&.411&.423&.400&.413&.422&.438&.410&.432&.411&.433&.398&.414&.424&.433\\
4&.526&.449&.409&.447&.449&.440&.534&.678&.556&.486&.428&.471&.464&.474&.567&.655&.557&.481&.425&.468&.467&.470&.572&.693\\
5&.582&.476&.411&.453&.453&.482&.547&.623&.605&.487&.434&.469&.459&.496&.565&.608&.621&.486&.438&.467&.470&.494&.586&.639\\
6&.351&.365&.330&.333&.368&.373&.342&.343&.335&.337&.322&.310&.264&.265&.324&.302&.376&.382&.359&.350&.353&.352&.365&.354\\
7&.466&.510&.442&.498&.493&.517&.481&.515&.408&.398&.397&.414&.418&.419&.404&.399&.472&.485&.450&.483&.495&.506&.476&.492\\
8&.293&.280&.280&.283&.276&.271&.284&.294&.208&.216&.231&.215&.248&.213&.194&.221&.262&.258&.271&.263&.276&.255&.251&.267\\
9&.210&.216&.196&.198&.194&.204&.195&.190&.210&.216&.198&.200&.196&.207&.198&.192&.211&.217&.198&.200&.196&.207&.197&.192\\
10&.007&.006&.008&.007&.009&.008&.007&.005&.051&.052&.052&.049&.052&.053&.051&.052&.037&.037&.038&.035&.038&.038&.037&.037\\
11&.617&.543&.532&.528&.552&.556&.693&.598&.628&.550&.549&.548&.565&.581&.677&.601&.640&.561&.553&.551&.568&.577&.714&.609\\
12&.507&.538&.527&.512&.574&.524&.668&.626&.554&.546&.575&.527&.597&.571&.662&.628&.551&.562&.565&.532&.597&.565&.691&.646\\
13&.730&.703&.700&.640&.703&.699&.724&.739&.667&.633&.615&.577&.590&.575&.628&.594&.719&.691&.682&.628&.676&.668&.703&.699\\
14&.416&.405&.380&.336&.530&.524&.411&.393&.427&.431&.400&.378&.239&.264&.411&.425&.439&.434&.410&.374&.453&.454&.430&.424\\
15&.838&.785&.836&.746&.812&.716&.834&.751&.794&.690&.794&.683&.793&.731&.785&.680&.841&.769&.839&.739&.827&.756&.833&.743\\
16&.713&.728&.716&.699&.692&.704&.715&.699&.689&.680&.720&.711&.647&.672&.682&.652&.715&.723&.727&.712&.689&.706&.714&.692\\
17&.258&.244&.259&.228&.266&.262&.256&.232&.213&.227&.211&.235&.220&.239&.211&.233&.255&.249&.254&.243&.262&.266&.252&.245\\
18&.522&.498&.446&.411&.469&.488&.472&.470&.516&.511&.447&.440&.387&.388&.472&.489&.537&.521&.465&.443&.459&.471&.491&.494\\
19&.525&.471&.516&.452&.542&.524&.526&.459&.516&.488&.488&.474&.540&.522&.511&.493&.554&.513&.538&.499&.573&.556&.553&.511\\
20&.800&.761&.783&.727&.774&.751&.806&.757&.693&.604&.706&.605&.714&.652&.707&.638&.772&.717&.762&.688&.760&.724&.779&.721\\
21&.703&.664&.606&.610&.648&.627&.742&.749&.701&.686&.630&.644&.652&.671&.642&.712&.725&.687&.630&.639&.666&.663&.748&.762\\
22&.541&.486&.455&.457&.489&.476&.826&.720&.560&.502&.467&.470&.497&.486&.785&.689&.564&.501&.468&.469&.501&.487&.841&.728\\
23&.366&.335&.346&.289&.380&.366&.323&.277&.354&.342&.332&.303&.191&.210&.273&.281&.369&.345&.348&.302&.324&.320&.313&.287\\
24&.261&.232&.165&.117&.200&.198&.190&.166&.485&.471&.433&.414&.404&.398&.438&.440&.428&.408&.361&.331&.361&.358&.373&.365\\
25&.568&.636&.543&.580&.551&.561&.698&.706&.673&.596&.765&.679&.713&.699&.486&.500&.664&.652&.710&.669&.682&.676&.651&.657\\
26&.694&.572&.677&.546&.682&.561&.701&.559&.687&.551&.667&.551&.656&.524&.681&.576&.701&.577&.684&.560&.684&.559&.704&.577\\
27&.578&.553&.562&.555&.591&.577&.562&.570&.575&.559&.575&.581&.590&.570&.575&.582&.596&.576&.586&.585&.609&.594&.586&.595\\
28&.345&.348&.334&.333&.350&.343&.349&.355&.357&.342&.344&.342&.295&.291&.357&.348&.396&.391&.384&.379&.369&.364&.397&.396\\
29&.509&.506&.419&.510&.531&.529&.502&.484&.335&.319&.458&.334&.240&.256&.351&.360&.460&.457&.470&.468&.445&.447&.464&.460\\
30&.456&.389&.392&.398&.422&.405&.436&.515&.281&.340&.299&.310&.288&.303&.309&.313&.406&.390&.375&.383&.389&.384&.409&.460\\

\end{tabular}
\end{sidewaystable}
}

{
\setlength\tabcolsep{1.4pt}
\begin{sidewaystable}
\centering\scriptsize
\caption{Base classifier comparison -- full results -- macro-averaged criteria and Hamming.}\label{table:FullResults2Base}
\begin{tabular}{c|cccccccc|cccccccc|cccccccc|cccccccc}
&\multicolumn{8}{c|}{Macro FDR}&\multicolumn{8}{c|}{Macro FNR}&\multicolumn{8}{c|}{Macro $F_1$}&\multicolumn{8}{c}{Hamming}\\
No.&1&2&3&4&5&6&7&8	&1&2&3&4&5&6&7&8	&1&2&3&4&5&6&7&8	&1&2&3&4&5&6&7&8\\
\hline
1 &.	569&.613&.563&.648&.583&.635&.564&.647&.760&.767&.811&.754&.798&.798&.769&.771&.715&.732&.768&.755&.769&.778&.718&.744	&.060&.064&.061&.086&.062&.065&.058&.068\\
2 &.	797&.879&.804&.909&.835&.901&.807&.856&.804&.857&.894&.919&.896&.926&.729&.817&.822&.883&.875&.918&.880&.924&.784&.853	&.168&.168&.124&.113&.135&.124&.214&.199\\
3 &.	717&.739&.708&.738&.713&.729&.697&.738&.777&.783&.777&.809&.789&.790&.767&.825&.768&.779&.761&.793&.768&.779&.752&.803	&.054&.055&.053&.058&.051&.054&.053&.053\\
4 &.	606&.917&.282&.941&.502&.866&.565&.870&.774&.970&.807&.982&.792&.974&.726&.775&.726&.962&.720&.976&.723&.959&.675&.838	&.127&.118&.099&.115&.111&.117&.126&.258\\
5 &.	724&.981&.520&.975&.584&.980&.682&.856&.764&.985&.812&.981&.796&.983&.701&.829&.758&.984&.757&.979&.753&.983&.697&.852	&.156&.122&.117&.121&.130&.123&.155&.228\\
6 &.	335&.345&.310&.327&.391&.394&.329&.341&.360&.353&.341&.320&.275&.279&.340&.319&.358&.356&.338&.332&.344&.349&.345&.338	&.207&.212&.195&.200&.224&.223&.202&.207\\
7 &.	763&.770&.751&.767&.759&.773&.759&.766&.729&.731&.739&.750&.716&.733&.728&.735&.758&.762&.757&.768&.752&.766&.755&.761	&.066&.070&.060&.066&.070&.076&.066&.071\\
8 &.	320&.342&.342&.357&.338&.308&.326&.347&.280&.295&.312&.327&.339&.300&.264&.307&.324&.332&.345&.362&.358&.325&.312&.344	&.248&.244&.255&.250&.248&.239&.240&.253\\
9 &.	618&.625&.547&.618&.554&.573&.566&.563&.693&.690&.692&.721&.647&.645&.692&.682&.672&.673&.654&.691&.621&.627&.656&.644	&.081&.084&.075&.078&.075&.079&.075&.073\\
10&.	250&.245&.255&.238&.242&.245&.246&.240&.246&.242&.250&.235&.239&.245&.242&.241&.248&.244&.252&.237&.241&.246&.244&.241	&.005&.005&.005&.005&.005&.005&.005&.005\\
11&.	847&.873&.831&.859&.844&.864&.829&.821&.818&.844&.853&.857&.870&.894&.745&.885&.847&.872&.852&.865&.864&.892&.805&.868	&.152&.137&.126&.120&.126&.122&.206&.145\\
12&.	820&.843&.859&.844&.806&.883&.819&.834&.839&.807&.881&.831&.852&.880&.811&.837&.844&.841&.877&.849&.841&.889&.829&.852	&.123&.146&.123&.128&.127&.127&.176&.171\\
13&.	837&.832&.800&.789&.795&.802&.817&.862&.846&.808&.806&.790&.789&.796&.803&.782&.847&.831&.812&.803&.812&.822&.832&.843	&.111&.123&.112&.110&.108&.114&.124&.156\\
14&.	291&.304&.275&.268&.502&.488&.323&.287&.447&.455&.425&.404&.269&.290&.439&.449&.383&.393&.361&.346&.445&.443&.389&.382	&.169&.175&.162&.158&.305&.293&.178&.170\\
15&.	911&.903&.906&.907&.907&.899&.908&.914&.899&.866&.890&.895&.902&.896&.893&.891&.914&.902&.912&.912&.907&.903&.909&.911	&.124&.136&.129&.114&.117&.110&.127&.118\\
16&.	653&.662&.668&.669&.651&.650&.658&.657&.658&.651&.672&.671&.646&.653&.658&.650&.662&.662&.674&.673&.653&.656&.662&.659	&.020&.025&.018&.018&.022&.022&.022&.026\\
17&.	382&.380&.381&.380&.383&.388&.383&.384&.360&.366&.358&.375&.361&.372&.360&.377&.377&.378&.375&.382&.378&.384&.377&.385	&.012&.012&.012&.011&.013&.012&.012&.011\\
18&.	407&.417&.342&.357&.477&.504&.398&.381&.536&.529&.469&.463&.408&.404&.490&.503&.483&.484&.417&.420&.453&.468&.454&.455	&.213&.219&.190&.196&.240&.256&.208&.205\\
19&.	513&.505&.536&.513&.505&.507&.524&.504&.591&.581&.567&.580&.610&.595&.581&.594&.578&.564&.574&.566&.585&.574&.577&.569	&.064&.062&.065&.063&.064&.063&.065&.062\\
20&.	849&.848&.814&.805&.835&.819&.816&.849&.784&.749&.811&.759&.834&.773&.806&.782&.837&.819&.821&.802&.843&.812&.832&.834	&.151&.172&.128&.151&.122&.138&.145&.161\\
21&.	809&.891&.857&.945&.853&.904&.830&.887&.810&.933&.919&.947&.898&.940&.692&.851&.826&.929&.911&.948&.888&.934&.789&.878	&.168&.135&.127&.121&.141&.123&.240&.209\\
22&.	825&.952&.826&.950&.828&.934&.882&.894&.927&.970&.964&.973&.945&.969&.756&.809&.909&.970&.945&.968&.923&.965&.843&.866	&.098&.087&.084&.085&.091&.087&.215&.206\\
23&.	185&.182&.172&.173&.346&.332&.249&.207&.356&.343&.335&.311&.202&.220&.280&.289&.286&.275&.268&.253&.290&.287&.272&.256	&.092&.090&.088&.085&.129&.122&.100&.090\\
24&.	151&.131&.106&.091&.202&.211&.141&.111&.583&.572&.543&.527&.516&.513&.542&.546&.446&.432&.400&.382&.405&.405&.407&.402	&.126&.122&.114&.110&.124&.125&.118&.115\\
25&.	680&.721&.733&.762&.679&.657&.744&.750&.787&.695&.882&.788&.828&.809&.563&.576&.774&.726&.881&.803&.809&.787&.678&.687	&.249&.308&.221&.274&.234&.240&.392&.396\\
26&.	516&.505&.522&.469&.535&.517&.523&.502&.633&.588&.620&.594&.609&.566&.621&.619&.598&.570&.591&.558&.589&.557&.596&.585	&.048&.051&.049&.046&.052&.051&.052&.048\\
27&.	598&.597&.586&.608&.574&.605&.571&.608&.644&.647&.681&.674&.701&.697&.673&.682&.658&.661&.676&.678&.671&.676&.669&.679	&.083&.080&.077&.079&.080&.081&.077&.084\\
28&.	449&.466&.434&.455&.438&.454&.423&.459&.545&.521&.506&.549&.464&.485&.550&.558&.528&.521&.498&.528&.474&.489&.520&.538	&.074&.075&.073&.073&.073&.073&.075&.076\\
29&.	532&.533&.491&.533&.534&.526&.534&.522&.374&.356&.531&.374&.285&.287&.393&.406&.473&.463&.543&.475&.458&.452&.477&.475	&.365&.368&.307&.378&.400&.395&.367&.353\\
30&.	625&.668&.572&.662&.588&.640&.608&.668&.494&.610&.519&.567&.483&.555&.507&.511&.573&.665&.558&.630&.546&.624&.569&.608	&.282&.239&.238&.248&.257&.248&.271&.335\\
\end{tabular}
\end{sidewaystable}
}

{
\setlength\tabcolsep{1.4pt}
\begin{sidewaystable}
\centering\scriptsize
\caption{Base classifier comparison -- full results -- micro-averaged criteria and zero one.}\label{table:FullResults3Base}

\begin{tabular}{c|cccccccc|cccccccc|cccccccc|cccccccc}
&\multicolumn{8}{c|}{Micro FDR}&\multicolumn{8}{c|}{Micro FNR}&\multicolumn{8}{c|}{Micro $F_1$}&\multicolumn{8}{c}{Zero one}\\
No.&1&2&3&4&5&6&7&8	&1&2&3&4&5&6&7&8	&1&2&3&4&5&6&7&8	&1&2&3&4&5&6&7&8\\
\hline
1 &.421&.478&.404&.636&.462&.504&.409&.553&.723&.696&.809&.667&.827&.812&.736&.683&.628&.621&.715&.668&.739&.733&.635&.630	&.809&.767&.861&.840&.887&.876&.811&.753\\
2 &.735&.722&.558&.491&.707&.584&.797&.817&.778&.844&.883&.923&.885&.923&.694&.789&.770&.826&.828&.868&.840&.880&.758&.816	&.759&.668&.575&.560&.617&.598&.852&.744\\
3 &.463&.487&.484&.551&.454&.500&.458&.479&.668&.682&.688&.741&.695&.706&.686&.745&.605&.617&.614&.680&.613&.636&.615&.663	&.529&.544&.522&.532&.510&.516&.539&.523\\
4 &.584&.593&.217&.444&.444&.551&.547&.858&.763&.966&.803&.977&.781&.968&.707&.756&.700&.937&.687&.956&.689&.942&.651&.822	&.612&.499&.455&.477&.497&.484&.638&.744\\
5 &.685&.593&.459&.450&.575&.595&.655&.851&.748&.983&.799&.978&.786&.981&.682&.824&.720&.968&.708&.959&.716&.964&.670&.845	&.707&.490&.488&.473&.517&.499&.685&.675\\
6 &.327&.340&.307&.326&.381&.380&.323&.340&.342&.337&.325&.304&.262&.264&.320&.302&.335&.339&.317&.316&.327&.328&.322&.322	&.734&.735&.703&.683&.733&.725&.705&.714\\
7 &.509&.538&.472&.516&.539&.567&.508&.537&.430&.419&.417&.445&.444&.444&.431&.423&.474&.486&.447&.484&.497&.516&.475&.490	&.907&.918&.914&.907&.929&.913&.920&.924\\
8 &.283&.271&.278&.277&.264&.262&.275&.283&.193&.208&.225&.212&.235&.204&.181&.208&.241&.241&.254&.247&.251&.236&.231&.248	&.820&.786&.793&.813&.821&.777&.771&.778\\
9 &.592&.606&.555&.568&.546&.578&.552&.524&.727&.705&.741&.758&.646&.642&.740&.732&.683&.677&.679&.700&.606&.616&.676&.660	&.218&.227&.204&.209&.204&.216&.204&.198\\
10&.008&.007&.009&.007&.008&.007&.008&.004&.101&.101&.104&.098&.100&.102&.101&.102&.057&.056&.059&.054&.056&.057&.057&.056	&.107&.109&.109&.104&.109&.107&.107&.107\\
11&.681&.632&.571&.558&.606&.595&.776&.719&.767&.791&.821&.815&.838&.850&.686&.851&.745&.759&.757&.742&.780&.789&.742&.812	&.725&.636&.623&.607&.614&.628&.847&.644\\
12&.483&.543&.445&.482&.539&.494&.697&.714&.769&.742&.822&.769&.822&.824&.753&.776&.686&.705&.738&.691&.749&.753&.739&.761	&.655&.665&.655&.614&.664&.657&.812&.743\\
13&.660&.676&.650&.620&.627&.635&.688&.747&.676&.639&.626&.588&.595&.578&.634&.594&.670&.665&.639&.612&.612&.614&.666&.693	&.859&.845&.848&.791&.864&.862&.875&.912\\
14&.295&.314&.279&.280&.569&.558&.331&.300&.452&.458&.431&.407&.281&.305&.443&.453&.384&.396&.364&.350&.461&.460&.392&.387	&.591&.564&.563&.509&.864&.854&.587&.550\\
15&.827&.809&.838&.757&.807&.761&.827&.766&.803&.717&.805&.713&.797&.749&.794&.707&.816&.772&.823&.737&.803&.756&.812&.740	&.985&.988&.986&.979&.975&.950&.986&.982\\
16&.701&.782&.615&.628&.725&.735&.746&.788&.818&.810&.852&.845&.775&.801&.812&.779&.782&.804&.797&.787&.754&.776&.790&.787	&.795&.808&.787&.763&.790&.792&.803&.795\\
17&.222&.203&.222&.178&.229&.206&.220&.184&.218&.230&.215&.241&.224&.241&.216&.240&.220&.217&.219&.211&.227&.224&.219&.213	&.409&.374&.409&.342&.414&.394&.406&.355\\
18&.409&.426&.357&.376&.485&.511&.405&.391&.544&.536&.474&.468&.418&.416&.498&.510&.486&.488&.423&.427&.454&.469&.456&.458	&.689&.662&.619&.589&.706&.728&.645&.617\\
19&.425&.410&.439&.429&.420&.419&.435&.413&.544&.519&.518&.508&.566&.550&.539&.529&.493&.471&.483&.472&.503&.493&.494&.477	&.814&.778&.814&.776&.821&.810&.815&.776\\
20&.745&.765&.711&.724&.687&.703&.738&.750&.696&.618&.714&.612&.718&.657&.708&.643&.734&.713&.714&.682&.708&.688&.736&.715	&.891&.897&.875&.872&.860&.855&.897&.894\\
21&.687&.670&.508&.534&.704&.546&.786&.827&.774&.904&.903&.926&.878&.916&.624&.797&.773&.864&.852&.874&.836&.865&.728&.818	&.833&.729&.687&.683&.722&.706&.925&.844\\
22&.744&.645&.497&.535&.699&.658&.876&.891&.919&.967&.957&.971&.940&.968&.744&.800&.880&.940&.922&.945&.901&.941&.833&.860	&.594&.514&.480&.479&.518&.500&.924&.774\\
23&.187&.186&.184&.188&.393&.369&.275&.226&.369&.355&.344&.318&.203&.222&.287&.297&.290&.281&.273&.259&.311&.304&.281&.264	&.444&.394&.419&.352&.595&.557&.431&.348\\
24&.159&.138&.113&.096&.217&.227&.148&.114&.582&.570&.542&.525&.514&.510&.542&.545&.442&.427&.396&.378&.401&.401&.404&.399	&.726&.712&.683&.662&.719&.726&.697&.684\\
25&.555&.649&.447&.615&.512&.531&.706&.710&.685&.607&.768&.684&.721&.706&.499&.508&.632&.629&.673&.653&.645&.639&.629&.636	&.954&.977&.947&.973&.952&.952&.994&.996\\
26&.392&.448&.412&.396&.472&.470&.454&.425&.690&.570&.672&.571&.659&.545&.683&.596&.591&.523&.580&.500&.587&.511&.605&.526	&.785&.689&.776&.651&.798&.688&.804&.656\\
27&.552&.524&.504&.524&.538&.538&.505&.561&.628&.614&.649&.627&.671&.654&.641&.623&.596&.577&.593&.585&.617&.607&.588&.598	&.741&.723&.716&.720&.737&.724&.720&.743\\
28&.360&.372&.355&.354&.372&.370&.364&.376&.390&.373&.378&.379&.322&.315&.387&.377&.375&.373&.367&.367&.348&.344&.376&.377	&.805&.797&.794&.777&.790&.780&.806&.808\\
29&.502&.504&.417&.515&.530&.527&.503&.488&.328&.316&.462&.331&.241&.249&.349&.359&.428&.426&.441&.438&.421&.421&.438&.432	&.995&.996&.991&.995&.999&1.00&.997&.994\\
30&.475&.398&.408&.416&.440&.422&.454&.537&.283&.352&.307&.322&.296&.315&.316&.323&.394&.378&.362&.375&.376&.374&.395&.450	&.948&.871&.900&.862&.910&.900&.918&.946\\

\end{tabular}
\end{sidewaystable}
}

{
\setlength\tabcolsep{0.4pt}
 \def\arraystretch{0.8}
\begin{table}
\centering\scriptsize
\caption{Naive Bayes -- Full results -- micro-averaged criteria. }\label{table:FullResults1}
\begin{tabular}{c|cccc|cccc|cccc}
&\multicolumn{4}{c|}{Micro FDR}&\multicolumn{4}{c|}{Micro FNR}&\multicolumn{4}{c}{Micro $F_1$}\\
No.&1&2&3&4&1&2&3&4&1&2&3&4\\
\hline
1&.523&.578&.636&.519&.926&.822&.923&.934&.873&.751&.873&.884\\
2&.513&.453&.437&.318&.956&.953&.958&.956&.922&.915&.923&.921\\
3&.469&.603&.541&.413&.636&.674&.633&.648&.573&.648&.596&.563\\
4&.462&.437&.409&.143&.845&.825&.848&.846&.761&.734&.760&.741\\
5&.550&.518&.546&.451&.839&.885&.868&.874&.764&.816&.797&.796\\
6&.392&.381&.389&.377&.279&.268&.277&.288&.340&.330&.338&.336\\
7&.742&.755&.777&.700&.511&.454&.517&.520&.663&.662&.695&.632\\
8&.256&.261&.262&.264&.210&.211&.217&.230&.234&.237&.241&.248\\
9&.628&.547&.640&.602&.673&.664&.691&.678&.659&.616&.674&.649\\
10&.215&.071&.469&.127&.443&.046&.419&.425&.351&.060&.450&.308\\
11&.415&.449&.458&.315&.927&.916&.938&.927&.873&.860&.891&.871\\
12&.303&.436&.355&.341&.874&.865&.867&.882&.790&.785&.785&.803\\
13&.586&.593&.592&.579&.639&.604&.620&.641&.615&.599&.607&.613\\
14&.574&.555&.574&.581&.308&.303&.305&.286&.473&.457&.472&.472\\
15&.692&.759&.724&.646&.926&.888&.907&.909&.881&.849&.863&.857\\
16&.971&.976&.984&.973&.654&.542&.511&.673&.947&.954&.969&.951\\
17&.335&.238&.415&.326&.681&.412&.686&.620&.571&.341&.596&.514\\
18&.482&.479&.489&.480&.455&.472&.458&.445&.469&.476&.474&.463\\
19&.334&.425&.340&.310&.728&.532&.728&.747&.614&.484&.615&.629\\
20&.553&.568&.577&.559&.844&.842&.828&.843&.771&.771&.756&.771\\
21&.448&.479&.530&.306&.958&.931&.933&.956&.923&.883&.883&.917\\
22&.641&.553&.522&.477&.954&.963&.960&.968&.919&.932&.927&.939\\
23&.402&.393&.403&.409&.198&.211&.195&.194&.315&.314&.315&.318\\
24&.262&.235&.254&.240&.504&.508&.503&.513&.407&.402&.404&.407\\
25&.410&.445&.421&.418&.781&.754&.773&.778&.681&.659&.674&.679\\
26&.235&.335&.239&.208&.841&.620&.834&.856&.738&.516&.728&.757\\
27&.359&.341&.348&.357&.838&.792&.834&.839&.742&.685&.736&.744\\
28&.389&.406&.390&.361&.316&.306&.312&.325&.355&.360&.354&.344\\
29&.500&.509&.503&.492&.333&.325&.350&.369&.429&.432&.437&.437\\
30&.364&.352&.367&.353&.344&.350&.346&.358&.355&.351&.357&.356\\
\end{tabular}
\end{table}
}

{
\setlength\tabcolsep{0.4pt}
 \def\arraystretch{0.8}
\begin{table}
\centering\scriptsize
\caption{Naive Bayes -- Full results -- Example based criteria, Hamming and zero-one.}\label{table:FullResults2}
\begin{tabular}{c|cccc|cccc|cccc|cccc|cccc}
 
&\multicolumn{4}{c|}{Hamming} & \multicolumn{4}{c|}{Zero-One}& \multicolumn{4}{c|}{EX FDR}& \multicolumn{4}{c|}{EX FNR}& \multicolumn{4}{c}{EX $F_1$}\\
 
No.&1&2&3&4&1&2&3&4&1&2&3&4&1&2&3&4&1&2&3&4\\
\hline
1&.064&.068&.068&.064&.936&.871&.936&.945&.902&.790&.902&.912&.910&.794&.907&.921&.912&.807&.911&.922\\
2&.112&.112&.111&.111&.553&.558&.555&.546&.540&.538&.536&.526&.527&.530&.527&.525&.536&.537&.534&.528\\
3&.052&.065&.058&.048&.496&.513&.510&.494&.349&.396&.361&.338&.376&.399&.381&.377&.380&.417&.390&.374\\
4&.113&.111&.110&.101&.480&.473&.475&.432&.435&.433&.434&.415&.425&.416&.420&.421&.438&.431&.433&.422\\
5&.125&.122&.124&.118&.473&.483&.481&.472&.439&.456&.456&.437&.421&.447&.441&.437&.440&.456&.455&.444\\
6&.232&.224&.231&.225&.709&.712&.711&.710&.399&.385&.395&.375&.282&.271&.283&.286&.368&.358&.367&.363\\
7&.123&.137&.141&.106&.871&.879&.867&.871&.495&.496&.513&.467&.482&.436&.487&.491&.546&.525&.560&.531\\
8&.234&.238&.241&.246&.789&.803&.814&.799&.264&.269&.272&.274&.218&.219&.226&.236&.252&.256&.260&.267\\
9&.088&.075&.088&.083&.217&.203&.218&.210&.200&.191&.203&.197&.190&.192&.193&.191&.198&.193&.201&.196\\
10&.028&.006&.045&.024&.560&.063&.554&.538&.514&.022&.523&.497&.511&.030&.494&.493&.518&.031&.521&.499\\
11&.113&.113&.112&.111&.565&.573&.573&.562&.546&.546&.554&.546&.536&.545&.544&.539&.544&.549&.552&.545\\
12&.115&.120&.115&.116&.622&.645&.625&.641&.560&.557&.554&.559&.585&.593&.578&.594&.579&.585&.572&.585\\
13&.097&.100&.099&.096&.839&.842&.846&.836&.704&.690&.699&.704&.632&.601&.615&.635&.690&.670&.681&.690\\
14&.307&.290&.307&.316&.874&.844&.872&.881&.542&.517&.541&.544&.266&.259&.263&.242&.466&.447&.464&.462\\
15&.077&.089&.082&.076&.960&.952&.962&.952&.862&.812&.833&.827&.915&.878&.901&.898&.901&.865&.886&.879\\
16&.196&.303&.481&.201&.838&.838&.985&.950&.795&.786&.943&.900&.549&.463&.570&.680&.787&.777&.932&.890\\
17&.023&.017&.026&.022&.777&.534&.796&.722&.657&.424&.687&.593&.686&.429&.698&.620&.683&.438&.705&.619\\
18&.238&.237&.242&.237&.684&.680&.693&.686&.467&.476&.477&.466&.427&.444&.431&.417&.472&.482&.479&.467\\
19&.063&.064&.063&.062&.844&.798&.848&.851&.644&.509&.644&.659&.701&.503&.702&.720&.695&.540&.696&.711\\
20&.094&.095&.096&.094&.875&.872&.870&.877&.847&.847&.833&.848&.840&.836&.825&.838&.847&.845&.834&.847\\
21&.118&.121&.120&.116&.683&.697&.692&.659&.638&.626&.620&.617&.646&.653&.624&.637&.648&.651&.632&.632\\
22&.087&.085&.085&.084&.482&.476&.486&.476&.463&.457&.468&.457&.467&.464&.474&.467&.471&.465&.475&.466\\
23&.132&.129&.132&.135&.604&.592&.601&.616&.382&.382&.378&.388&.188&.201&.185&.184&.323&.327&.319&.326\\
24&.130&.126&.128&.127&.740&.730&.739&.732&.216&.203&.212&.213&.390&.396&.389&.401&.363&.360&.359&.367\\
25&.216&.220&.216&.217&.952&.942&.952&.951&.553&.512&.546&.547&.782&.753&.773&.778&.724&.693&.716&.719\\
26&.048&.044&.048&.048&.855&.662&.849&.870&.818&.579&.811&.834&.833&.597&.825&.849&.829&.597&.822&.845\\
27&.070&.068&.070&.070&.724&.701&.724&.722&.637&.604&.641&.636&.667&.630&.667&.666&.662&.628&.663&.660\\
28&.076&.079&.076&.071&.772&.777&.769&.772&.341&.354&.342&.330&.292&.284&.290&.300&.365&.368&.364&.360\\
29&.363&.372&.365&.355&.998&1.000&.998&.999&.505&.512&.508&.497&.330&.323&.349&.365&.454&.458&.463&.462\\
30&.218&.213&.220&.214&.839&.835&.837&.837&.353&.345&.358&.344&.334&.342&.336&.347&.370&.369&.373&.373\\
  \hline
\end{tabular}
\end{table}
}

{
\setlength\tabcolsep{0.4pt}
 \def\arraystretch{0.8}
\begin{table}
\centering\scriptsize
\caption{Naive Bayes -- Full results -- Macro averaged criteria.}\label{table:FullResults3}
\begin{tabular}{c|cccc|cccc|cccc}
 
& \multicolumn{4}{c|}{Macro FDR} &\multicolumn{4}{c|}{Macro FNR}&\multicolumn{4}{c}{Macro $F_1$} \\
 
No.&1&2&3&4&1&2&3&4&1&2&3&4\\
\hline
1  &.650&.667&.715&.683&.904&.849&.929&.936&.886&.832&.914&.916 \\
2  &.905&.917&.922&.883&.948&.945&.952&.952&.941&.939&.943&.937 \\
3  &.629&.705&.661&.587&.716&.753&.710&.709&.702&.749&.707&.682 \\
4  &.577&.525&.569&.440&.861&.840&.865&.862&.804&.781&.807&.798 \\
5  &.645&.664&.696&.596&.843&.888&.871&.878&.794&.843&.829&.824 \\
6  &.406&.391&.403&.393&.292&.282&.289&.302&.366&.352&.361&.364 \\
7  &.860&.874&.882&.861&.766&.729&.762&.808&.855&.851&.870&.866 \\
8  &.296&.302&.298&.310&.302&.304&.306&.326&.318&.322&.323&.339 \\
9  &.728&.534&.750&.689&.689&.629&.728&.707&.739&.601&.765&.717 \\
10 &.285&.172&.591&.308&.353&.139&.502&.380&.351&.162&.579&.376 \\
11 &.885&.875&.902&.877&.926&.922&.936&.926&.917&.909&.928&.912 \\
12 &.831&.855&.848&.858&.908&.907&.904&.918&.888&.892&.889&.900 \\
13 &.765&.759&.758&.761&.827&.814&.818&.827&.824&.814&.820&.826 \\
14 &.491&.483&.495&.499&.292&.288&.288&.272&.454&.440&.453&.452 \\
15 &.920&.908&.903&.909&.979&.954&.962&.973&.972&.949&.958&.968 \\
16 &.929&.977&.985&.977&.768&.719&.700&.836&.913&.961&.972&.962 \\
17 &.508&.475&.591&.479&.544&.500&.616&.521&.533&.496&.608&.511 \\
18 &.469&.460&.482&.463&.441&.459&.443&.431&.475&.476&.483&.467 \\
19 &.442&.518&.445&.434&.783&.580&.783&.804&.717&.566&.718&.736 \\
20 &.816&.803&.824&.821&.905&.901&.896&.901&.884&.877&.878&.880 \\
21 &.895&.913&.933&.916&.965&.951&.941&.963&.951&.945&.942&.955 \\
22 &.772&.807&.792&.824&.955&.963&.959&.968&.930&.948&.947&.957 \\
23 &.346&.343&.347&.352&.198&.211&.195&.194&.290&.292&.289&.292 \\
24 &.245&.219&.239&.225&.508&.511&.507&.518&.415&.406&.410&.415 \\
25 &.781&.710&.724&.785&.894&.876&.887&.893&.894&.878&.885&.895 \\
26 &.473&.475&.491&.472&.777&.652&.778&.782&.733&.613&.733&.740 \\
27 &.683&.568&.629&.671&.898&.837&.890&.897&.853&.783&.840&.850 \\
28 &.511&.527&.514&.447&.462&.453&.448&.486&.505&.510&.502&.487 \\
29 &.497&.513&.495&.490&.380&.367&.394&.422&.479&.478&.488&.505 \\
30 &.544&.537&.554&.530&.570&.588&.575&.581&.580&.595&.587&.583 \\
  
\end{tabular}
\end{table}
}

{
\setlength\tabcolsep{0.4pt}
 \def\arraystretch{0.8}
\begin{table}
\centering\scriptsize
\caption{Nearest Neighbours -- Full results -- micro-averaged criteria. }\label{table:FullResults1KNN}
\begin{tabular}{c|cccc|cccc|cccc}
&\multicolumn{4}{c|}{Micro FDR}&\multicolumn{4}{c|}{Micro FNR}&\multicolumn{4}{c}{Micro $F_1$}\\
No.&1&2&3&4&1&2&3&4&1&2&3&4\\
\hline
1  &.387&.308&.535&.520&.760&.757&.742&.715&.656&.640&.668&.643\\
2  &.467&.856&.500&.442&.985&.937&.983&.977&.971&.913&.968&.955\\
3  &.203&.256&.173&.218&.815&.819&.825&.810&.702&.710&.715&.697\\
4  &.405&.867&.450&.497&.985&.904&.986&.987&.972&.889&.973&.975\\
5  &.403&.851&.472&.484&.980&.900&.983&.983&.961&.881&.967&.967\\
6  &.314&.340&.322&.322&.301&.306&.302&.295&.308&.324&.313&.309\\
7  &.416&.394&.415&.387&.514&.514&.560&.610&.470&.461&.498&.523\\
8  &.259&.281&.271&.273&.226&.215&.208&.191&.243&.250&.242&.235\\
9  &.444&.548&.501&.486&.862&.758&.875&.862&.787&.686&.806&.789\\
10 &.004&.003&.003&.004&.191&.111&.180&.206&.108&.060&.100&.117\\
11 &.337&.689&.440&.443&.967&.900&.964&.961&.939&.851&.934&.930\\
12 &.264&.585&.280&.390&.906&.861&.897&.905&.836&.803&.822&.840\\
13 &.525&.546&.549&.546&.770&.788&.780&.776&.691&.711&.704&.701\\
14 &.309&.312&.353&.323&.463&.437&.474&.461&.396&.381&.420&.400\\
15 &.665&.631&.703&.641&.945&.860&.869&.882&.907&.798&.818&.824\\
16 &.679&.641&.700&.650&.852&.849&.842&.858&.799&.790&.794&.799\\
17 &.189&.187&.197&.195&.240&.239&.272&.262&.215&.214&.237&.230\\
18 &.368&.392&.416&.375&.564&.519&.571&.568&.484&.463&.505&.490\\
19 &.359&.368&.436&.440&.567&.557&.564&.560&.483&.479&.508&.507\\
20 &.543&.610&.518&.545&.879&.871&.839&.818&.810&.808&.760&.741\\
21 &.417&.826&.373&.391&.962&.922&.950&.947&.930&.898&.908&.903\\
22 &.479&.880&.534&.542&.978&.925&.979&.978&.958&.908&.959&.957\\
23 &.231&.239&.266&.272&.309&.291&.310&.320&.272&.266&.289&.296\\
24 &.160&.191&.178&.180&.541&.540&.539&.546&.407&.414&.410&.416\\
25 &.483&.516&.500&.476&.774&.756&.768&.772&.685&.676&.684&.683\\
26 &.347&.320&.520&.574&.665&.639&.644&.658&.561&.529&.592&.621\\
27 &.350&.356&.364&.340&.737&.712&.735&.770&.627&.603&.627&.660\\
28 &.319&.327&.323&.335&.435&.409&.427&.401&.383&.371&.380&.370\\
29 &.431&.439&.433&.424&.449&.436&.472&.468&.440&.438&.454&.447\\
30 &.343&.390&.381&.372&.452&.455&.420&.381&.403&.425&.402&.376\\
\end{tabular}
\end{table}
}

{
\setlength\tabcolsep{0.4pt}
 \def\arraystretch{0.8}
\begin{table}
\centering\scriptsize
\caption{Nearest Neighbours -- Full results -- Example based, Hamming and zero-one criteria.}\label{table:FullResults2KNN}
\begin{tabular}{c|cccc|cccc|cccc|cccc|cccc}
 
&\multicolumn{4}{c|}{Hamming} & \multicolumn{4}{c|}{Zero-One}& \multicolumn{4}{c|}{EX FDR}& \multicolumn{4}{c|}{EX FNR}& \multicolumn{4}{c}{EX $F_1$}\\
 
No.&1&2&3&4&1&2&3&4&1&2&3&4&1&2&3&4&1&2&3&4\\
\hline
1  &.058&.055&.066&.065&.777&.763&.756&.732&.658&.648&.630&.591&.715&.703&.685&.654&.700&.689&.672&.639\\
2  &.112&.152&.113&.112&.563&.597&.560&.560&.553&.563&.553&.548&.558&.559&.557&.554&.556&.572&.555&.552\\
3  &.046&.047&.046&.046&.498&.496&.495&.501&.392&.397&.398&.391&.446&.447&.449&.447&.431&.434&.435&.432\\
4  &.115&.175&.115&.115&.473&.596&.475&.477&.459&.565&.461&.465&.469&.568&.471&.473&.467&.576&.469&.472\\
5  &.120&.181&.121&.121&.472&.569&.470&.471&.454&.540&.460&.461&.469&.538&.468&.469&.467&.551&.467&.468\\
6  &.194&.208&.198&.197&.662&.712&.704&.675&.321&.341&.316&.329&.302&.307&.299&.296&.340&.356&.341&.341\\
7  &.055&.053&.056&.055&.978&.876&.905&.887&.445&.402&.430&.393&.552&.483&.544&.564&.525&.472&.518&.517\\
8  &.242&.254&.245&.241&.790&.784&.801&.790&.268&.295&.285&.280&.228&.227&.215&.195&.258&.271&.260&.250\\
9  &.069&.073&.070&.069&.192&.200&.194&.192&.184&.191&.187&.184&.186&.194&.189&.185&.186&.193&.189&.186\\
10 &.009&.005&.008&.010&.153&.109&.151&.156&.022&.004&.026&.026&.092&.055&.087&.098&.071&.039&.068&.078\\
11 &.112&.128&.113&.114&.567&.599&.573&.578&.554&.553&.561&.560&.557&.570&.565&.566&.557&.567&.564&.565\\
12 &.117&.134&.117&.121&.638&.671&.628&.641&.544&.575&.526&.556&.592&.622&.581&.602&.578&.609&.565&.588\\
13 &.087&.088&.089&.088&.825&.835&.830&.825&.750&.769&.757&.757&.767&.784&.778&.772&.766&.783&.775&.771\\
14 &.174&.171&.188&.178&.544&.549&.569&.548&.385&.385&.401&.387&.429&.408&.444&.430&.422&.413&.439&.424\\
15 &.076&.078&.084&.077&.974&.933&.939&.942&.894&.745&.756&.774&.939&.846&.855&.867&.927&.819&.828&.842\\
16 &.018&.018&.019&.018&.762&.763&.774&.772&.701&.700&.703&.711&.717&.714&.705&.722&.716&.714&.713&.724\\
17 &.012&.012&.013&.012&.361&.353&.372&.378&.237&.232&.256&.264&.235&.231&.266&.264&.248&.244&.272&.275\\
18 &.203&.205&.217&.205&.629&.632&.639&.637&.486&.477&.503&.489&.537&.500&.545&.545&.524&.504&.537&.530\\
19 &.059&.059&.065&.066&.762&.766&.766&.772&.462&.462&.463&.465&.528&.520&.526&.523&.527&.523&.527&.527\\
20 &.092&.096&.092&.093&.890&.875&.847&.834&.870&.864&.829&.808&.878&.865&.834&.816&.876&.866&.833&.814\\
21 &.118&.162&.117&.117&.671&.720&.659&.666&.638&.677&.614&.617&.657&.684&.640&.643&.652&.693&.632&.635\\
22 &.084&.125&.085&.085&.474&.547&.482&.480&.454&.523&.462&.460&.468&.525&.475&.473&.466&.532&.473&.471\\
23 &.092&.092&.100&.103&.349&.353&.367&.375&.280&.278&.297&.309&.292&.276&.295&.305&.293&.286&.304&.315\\
24 &.120&.124&.122&.123&.693&.711&.703&.706&.182&.196&.193&.202&.432&.436&.433&.442&.368&.377&.374&.381\\
25 &.227&.235&.231&.226&.954&.953&.947&.955&.601&.593&.600&.593&.772&.750&.762&.766&.730&.716&.724&.725\\
26 &.046&.043&.055&.060&.693&.666&.687&.737&.615&.590&.598&.628&.647&.617&.626&.648&.638&.611&.621&.649\\
27 &.067&.066&.067&.067&.682&.673&.680&.696&.547&.532&.547&.576&.612&.596&.609&.633&.593&.578&.591&.617\\
28 &.071&.070&.071&.071&.778&.775&.775&.774&.324&.321&.319&.324&.404&.377&.395&.367&.408&.392&.401&.387\\
29 &.314&.318&.317&.312&.985&.990&.993&.993&.424&.436&.434&.427&.446&.435&.464&.459&.467&.469&.481&.476\\
30 &.224&.243&.236&.226&.857&.894&.869&.848&.337&.384&.379&.368&.456&.449&.416&.370&.427&.448&.419&.389\\
\end{tabular}
\end{table}
}

{
\setlength\tabcolsep{1.4pt}
\begin{table}
\centering\scriptsize
\caption{Nearest Neighbours -- Full results -- Macro-averaged criteria.}\label{table:FullResults3KNN}
\begin{tabular}{c|cccc|cccc|cccc}
 
& \multicolumn{4}{c|}{Macro FDR} &\multicolumn{4}{c|}{Macro FNR}&\multicolumn{4}{c}{Macro $F_1$}\\
 
No.&1&2&3&4&1&2&3&4&1&2&3&4\\
\hline
1  &.565&.550&.692&.579&.823&.822&.823&.803&.772&.768&.791&.766 \\
2  &.958&.907&.956&.956&.985&.945&.984&.978&.979&.934&.977&.971 \\
3  &.736&.723&.730&.726&.871&.871&.876&.867&.838&.835&.842&.833 \\
4  &.972&.873&.973&.976&.987&.914&.988&.989&.983&.900&.983&.985 \\
5  &.973&.858&.976&.977&.982&.903&.985&.985&.979&.887&.982&.982 \\
6  &.308&.344&.325&.314&.321&.324&.322&.315&.327&.343&.334&.325 \\
7  &.746&.748&.744&.744&.789&.787&.792&.796&.783&.782&.781&.792 \\
8  &.373&.343&.375&.373&.361&.319&.332&.313&.386&.348&.370&.364 \\
9  &.593&.566&.627&.595&.761&.699&.773&.761&.728&.660&.744&.728 \\
10 &.324&.254&.317&.358&.344&.258&.333&.359&.338&.256&.328&.359 \\
11 &.898&.829&.915&.906&.971&.928&.972&.967&.958&.906&.960&.954 \\
12 &.893&.840&.898&.922&.947&.910&.945&.950&.933&.895&.934&.945 \\
13 &.766&.782&.805&.785&.913&.918&.916&.916&.894&.900&.899&.900 \\
14 &.301&.300&.340&.314&.461&.434&.469&.459&.394&.377&.417&.398 \\
15 &.940&.930&.941&.956&.976&.961&.958&.966&.974&.960&.956&.967 \\
16 &.649&.649&.649&.652&.671&.671&.667&.673&.669&.668&.668&.670 \\
17 &.393&.387&.395&.396&.378&.375&.389&.382&.390&.385&.397&.394 \\
18 &.368&.385&.410&.380&.556&.511&.560&.552&.487&.460&.510&.505 \\
19 &.479&.471&.549&.499&.625&.616&.623&.616&.582&.574&.601&.590 \\
20 &.849&.829&.835&.821&.943&.932&.928&.916&.924&.911&.908&.897 \\
21 &.939&.892&.938&.944&.972&.942&.963&.960&.964&.930&.956&.955 \\
22 &.961&.877&.961&.964&.981&.929&.981&.980&.975&.913&.976&.975 \\
23 &.206&.217&.240&.246&.299&.283&.302&.310&.265&.258&.281&.288 \\
24 &.157&.183&.176&.179&.543&.541&.541&.548&.411&.417&.414&.421 \\
25 &.709&.716&.691&.680&.872&.857&.865&.870&.856&.840&.850&.853 \\
26 &.474&.467&.574&.476&.659&.650&.650&.657&.615&.608&.633&.631 \\
27 &.571&.516&.621&.626&.790&.758&.788&.818&.750&.715&.752&.780 \\
28 &.422&.414&.410&.415&.639&.607&.632&.606&.592&.563&.579&.561 \\
29 &.480&.480&.483&.477&.513&.497&.538&.532&.512&.501&.531&.525 \\
30 &.624&.640&.645&.691&.683&.676&.669&.643&.690&.670&.697&.692 \\
\end{tabular}
\end{table}
}

\end{document}